\documentclass[11pt]{article}

\usepackage[final]{acl}

\usepackage{times}
\usepackage{latexsym}

\usepackage[T1]{fontenc}
\usepackage[utf8]{inputenc}

\usepackage{microtype}

\usepackage{inconsolata}

\usepackage{graphicx}
\usepackage{amsmath}
\usepackage{amssymb}
\usepackage{booktabs}

\usepackage{cleveref}
\usepackage[most]{tcolorbox}
\usepackage{multirow}
\usepackage{tikz}
\usepackage{listings}

\newcommand{\mc}[1]{\mathcal{#1}}

\newcommand{\ours}{\textbf{CLEAR}}
\newcommand{\footnoteurl}[1]{\footnote{\;\url{#1}}}
\newcommand{\cmd}[1]{{\ttfamily #1}}

\DeclareMathOperator*{\argmax}{arg\,max}

\usetikzlibrary{
  arrows.meta,
  positioning,
  fit,
  calc,
  shapes.symbols,
  shapes.geometric,
  shadows.blur
}
\title{\ours: Context Augmentation from Contrastive Learning of Experience via Agentic Reflection}

\author{Linbo Liu \quad Guande Wu \quad Han Ding \quad Yawei Wang \quad Qiang Zhou \\ \bf Yuzhe Lu \quad Zhichao Xu \quad Huan Song \quad Panpan Xu \quad Lin Lee Cheong \\
        AWS AI\\
    \small{
    \textbf{Correspondence:} \href{mailto:linbol@amazon.com}{linbol@amazon.com}
}
}
\begin{document}
\maketitle
\begin{abstract}
Large language model agents rely on effective model context to obtain task-relevant information for decision-making. Many existing context engineering approaches primarily rely on the context generated from the past experience and retrieval mechanisms that reuse these context. However, retrieved context from past tasks must be adapted by the execution agent to fit new situations, placing additional reasoning burden on the underlying LLM. To address this limitation, we propose a generative context augmentation framework using \textbf{C}ontrastive \textbf{L}earning of \textbf{E}xperience via \textbf{A}gentic \textbf{R}eflection (\ours). \ours\ first employs a reflection agent to perform contrastive analysis over past execution trajectories and summarize useful context for each observed task. These summaries are then used as supervised fine-tuning data to train a context augmentation model (CAM). Then we further optimize CAM using reinforcement learning, where the reward signal is obtained by running the task execution agent. By learning to {generate} task-specific knowledge rather than {retrieve} knowledge from the past, CAM produces context that is better tailored to the current task. We conduct comprehensive evaluations on the AppWorld and WebShop benchmarks. Experimental results show that \ours\ consistently outperforms strong baselines. It improves task completion rate from 72.62\% to 81.15\% on AppWorld test set and averaged reward from 0.68 to 0.74 on a subset of WebShop, compared with baseline agent. Our code is publicly available at \url{https://github.com/awslabs/CLEAR}.
\end{abstract}

\section{Introduction}

Large language models (LLMs) have become increasingly powerful as their parameter scale continues to grow~\citep{xi2025rise, kaplan2020scaling, wei2022emergent, ding2024reasoning}. 
Recent works have demonstrated that LLMs can act as agents in sequential decision-making settings and achieve strong performance across a variety of tasks~\citep{yao2023react, talebirad2023multi, wang2024executable, xia2025live, black2024pi_0, hu2025qualityflow}. 
Despite these advances, LLMs still rely primarily on parametric knowledge stored in their model weights when performing reasoning, which can be outdated, incomplete, or insufficient for complex, knowledge-intensive tasks~\citep{lewis2020retrieval, gao2023retrieval, suzgun2025dynamic}. Effective integration of external knowledge and task-relevant context remains a key challenge in improving agent decision-making capabilities. 

Retrieval augmented generation (RAG)~\citep{lewis2020retrieval} and other context engineering techniques~\citep{mei2025survey} are proposed to bridge the gap between parametric knowledge and context integration. However, typical RAG systems face several practical challenges, including designing effective knowledge base indexing strategies~\citep{huang2025ket}, query rewriting~\citep{ma2023query,peng2024large,xu2025rethinkingonpolicyoptimizationquery}, and devising reliable retrieval pipelines~\citep{jin2025search,shao2023enhancing,yu2022generate,xu2026asurveyofmodelarchitectures}. Moreover, their performance heavily depends on the quality and relevance of the underlying knowledge base. 
Several recent works on prompt optimization attempt to address these limitations. For example,
Agentic Context Engineering (ACE)~\citep{zhang2025agenticcontextengineeringevolving} and Dynamic Cheatsheet~\citep{suzgun2025dynamic} learn instructions from the past experience of LLM agents and reuse them to assist decision-making on future tasks. GEPA~\citep{agrawal2025gepa} proposes an iterative prompt optimization framework using pareto-based candidate selection. \cite{li2024learning} trains a prompt rewriter to generate the best prompt.
However, these universally learned guidance or optimized prompts are typically static and general-purpose, rather than tailored to specific future task instance. 
Consequently, the execution agent must reason about how to adapt them to the current task. 
This requirement can become problematic when the underlying LLM has limited reasoning capability, or when the future task differs substantially from previous ones, in which case the stored guidance and optimized prompts may be only weakly relevant. See \Cref{app:rag} for a detailed discussion.
% \zx{this paragraph currently reads as the motivation is to improve the performance of potentially out-of-distribution future task, but the same point is not directly addressed by the subsequent paragraph/proposed approach}

To address this limitation, we propose a context augmentation framework using \textbf{C}ontrastive \textbf{L}earning of \textbf{E}xperience via \textbf{A}gentic \textbf{R}eflection (\ours). 
\ours\ first employs a reflection agent to perform contrastive analysis over past experience replays and summarize useful context for each task. 
The resulting context is then used as supervised fine-tuning (SFT) data to train a context augmentation model (CAM). 
After the SFT stage, we further optimize the CAM using reinforcement learning (RL), where the reward signal is obtained by executing the task execution agent (see \Cref{fig:clear-workflow}). 
Further, the choice of the CAM can be lightweight and agnostic to the choice of the expensive execution agent, adding negligible overhead to the overall system.

% To improve efficiency, the CAM is typically selected to be a smaller model than the execution agent.
% \zx{essentially, the proposed framework highlights the importance of ``contrastive learning'', therefore, would it be better to mention this is a critical limitation of prior works to not consider failed trajectories?} 
% \zx{To double down the efficiency argument, the last sentence can be alternative phrased as ``further, the choice of the context optimization model can be lightweight and agnostic to the choice of the expensive execution agent, improving the efficiency of the overall system.}

\begin{figure*}
    \centering
    \includegraphics[width=1.7\columnwidth]{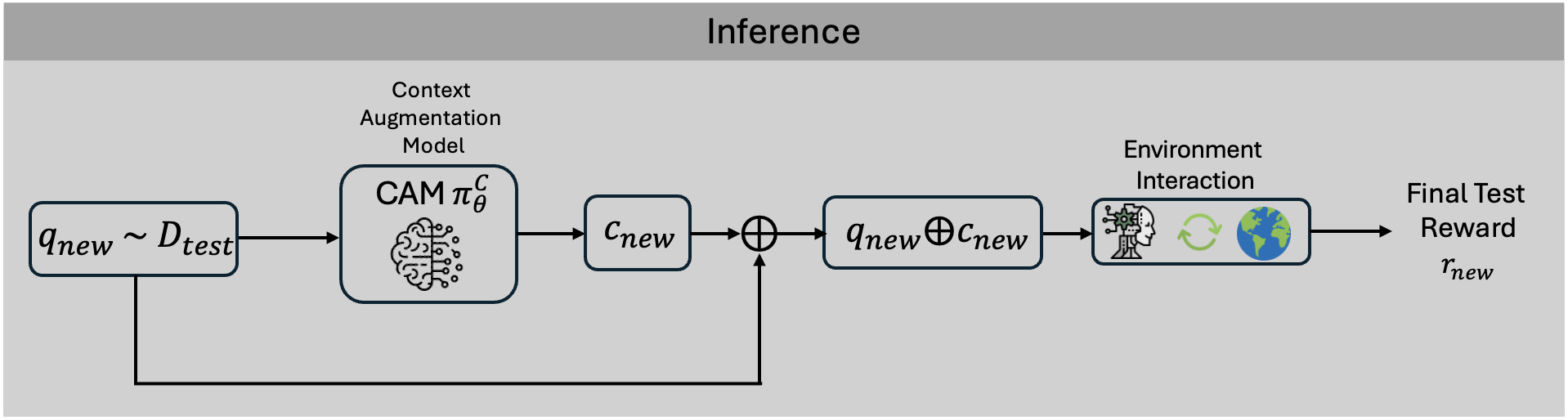}
    \caption{During inference, a new task $q_{\text{new}}$ is sampled from $\mc{D}_{\text{test}}$ and is passed into $\pi^C_\theta$ to generate $c_{\text{new}}\sim\pi^C_\theta(q_{\text{new}})$. The auxiliary context $c_{\text{new}}$ is appended to $q_{\text{new}}$ and the execution agent $\pi^E$ starts with $q_{\text{new}}\oplus c_{\text{new}}$.}
    \label{fig:inference}
\end{figure*}

The trained CAM provides additional context that is useful for solving future tasks, which will be integrated into the prompt of the task execution agent, as shown in \Cref{fig:inference}. 
Importantly, \ours\ does not require parametric training of the underlying LLM for execution agents, which are often proprietary models with no access to their weights. 
Instead, \ours\ only requires training the smaller CAM. 
As a result, \ours\ is a unified framework that can be applied to LLM agents built on either proprietary or open-source foundation models.

Our \ours\ framework has the following contributions:
\begin{itemize}
    \item We propose a CAM that generates additional context to improve the performance of LLM agents. CAM integrates task-relevant context into the prompt of the execution agent, avoiding any modification of the underlying LLM weights and making the framework broadly applicable across different agentic systems.

    \item For CAM training data generation, we introduce an agentic reflection mechanism that performs contrastive learning over past execution trajectories. By systematically analyzing multiple trajectories, the reflection agent extracts high-quality instructions as SFT training data for CAM.

    \item We design a two-stage training pipeline (SFT + RL) to train the CAM. In particular, we build a novel RL training framework that couples the CAM with the execution agent to generate rollouts, while only updating CAM's parameters.

    \item We evaluate \ours\ against diverse context engineering methods, including RAG and ACE,
    % and \textsc{GenRead}~\cite{yu2022generate}, 
    across multiple benchmarks. \ours\ consistently outperforms all baselines on every evaluated dataset.
\end{itemize}
\section{Related Work}

\paragraph{LLM Agents.}
LLM agents extend foundation models into autonomous, goal-directed systems by augmenting them with planning, memory, tool use, and action modules \citep{wang2024survey, xi2025rise}. LLM agents are built upon the foundational reasoning capabilities, demonstrated by works such as CoT~\citep{wei2022chain}, ReAct~\citep{yao2023react}, ToT~\citep{yao2023tree}, and DoT~\citep{lingam2025enhancing}. When equipped with tools, LLMs can learn to invoke external APIs to overcome inherent limitations in calculation, retrieval, and real-world interaction~\citep{schick2023toolformer, patil2024gorilla, qian2025toolrl}. Later, Anthropic's Model Context Protocol (MCP) \citep{anthropic2025mcp} proposed a standardized open protocol for connecting LLM agents to external tools and data sources, addressing the fragmentation of tool integration interfaces.
Browser-use and terminal-use agents have also matured, with Operator \citep{openai2025operator} and Claude Code\footnoteurl{https://github.com/anthropics/claude-code} demonstrating agents that autonomously navigate web browsers and terminal environments to complete real-world tasks.

On the multi-agent front, several frameworks have demonstrated that collaboration among specialized LLM agents can tackle complex tasks more effectively than single agents. {CAMEL} \citep{li2023camel} explored role-playing-based cooperative communication, while {MetaGPT} \citep{hong2024metagpt} and {ChatDev} \citep{qian2024chatdev} organized agents into software-engineering teams following structured workflows. {AutoGen} \citep{wu2024autogen} provided a general-purpose multi-agent conversation framework with human-in-the-loop support. 
% For orchestrating diverse AI models, \textbf{HuggingGPT} \cite{shen2023hugginggpt} used an LLM as a controller to coordinate task-specific models. The release of OpenAI Agents SDK \cite{openai2025agentssdk} and Manus \cite{manus2025} signaled a shift toward production-grade agentic platforms, with Manus demonstrating a general-purpose autonomous agent capable of end-to-end task completion across web browsing, coding, and file management. Evaluation has been facilitated by benchmarks such as AgentBench \cite{liu2023agentbench}, WebArena \cite{zhou2024webarena}, SWE-bench \cite{jimenez2024swebench}, and $\tau$-bench \cite{yao2024taubench}, all revealing a persistent gap between agent and human performance. Despite rapid progress, key challenges remain in reliability, long-horizon planning, error accumulation, and safety \cite{ruan2024toolemu}, motivating continued research into more robust and trustworthy agentic systems.

\paragraph{Contrastive Learning.} 
% \red{@Zhichao Xu}
% Traditional contrastive learning -> Recent LLM contrastive learning -> Our contrastive learning.
Contrastive signals are widely used to learn robust representations by explicitly comparing informative alternatives~\citep{gutmann10noisecontrastiveestimation,ma-collins-2018-noise,oord2019representationlearningcontrastivepredictive,zhang-stratos-2021-understanding,xu-etal-2025-distillation}. 
In LLM-agent settings, a closely related idea appears in reflection-based methods that learn from behavioral differences across trials, especially between successful and failed executions~\citep{shinn2023reflexion, wang2023voyager, Yu2026SelfConsolidation, Forouzandeh2025MACLA, Allard2026ERL}. 
Our work applies this principle in a practical agent-training pipeline: we contrast multiple rollouts for the same task and distill reusable strategy-level context through an agentic reflector. 
The novelty is mainly in this task-level adaptation and integration with context augmentation, rather than in a new contrastive objective itself.

\paragraph{Context Engineering.}
% \red{@Zhichao Xu}
% \begin{itemize}
%     \item Existing context engineering work. 
%     \item RAG.
%     \item Mention ACE~\citep{zhang2025agenticcontextengineeringevolving} and GEPA~\citep{agrawal2025gepa}.
%     \item Discuss how \ours\ is different.
% \end{itemize}
Context engineering aims to provide LLM agents with task-relevant information at inference time. 
Retrieval-based methods such as RAG~\citep{lewis2020retrieval, gao2023retrieval} improve factual grounding by retrieving external evidence, and generate-then-read variants further combine parametric generation with retrieval to improve coverage~\citep{yu2022generate}. 
Recent agent-centric approaches, including ACE~\citep{zhang2025agenticcontextengineeringevolving} and MIPRO~\citep{opsahl2024optimizing}, maintain evolving playbooks distilled from prior executions, while GEPA~\citep{agrawal2025gepa} optimizes prompts through reflective evolution. 
Compared with these methods, \ours\ learns a dedicated context augmentation model that maps a new task directly to actionable context via SFT+RL, rather than relying on nearest-neighbor retrieval or purely prompt-level updates. 
This shifts more adaptation into a trainable model and reduces the burden on the execution agent to reinterpret retrieved past experience.

\paragraph{LLM Fine-Tuning.}
The dominant paradigm for post-training large language models follows a two-stage pipeline: SFT on curated demonstrations, followed by RL to further align model behavior with desired objectives~\citep{ouyang2022training, bai2022training}. The RL stage has been realized through various algorithms, including PPO~\citep{schulman2017proximal}, which optimizes a clipped surrogate objective with a learned value function; DPO~\citep{rafailov2023direct}, which bypasses explicit reward modeling by directly optimizing on preference pairs; GRPO~\citep{shao2024deepseekmath}, which computes group-relative advantages to eliminate the critic model entirely; and many others \citep{zhang2021sample, ahmadian2024back, yu2025dapo, yue2025vapo, zheng2025group}. \ours\ follows the same two-stage training paradigm of SFT followed by RL and is agnostic to the choice of RL algorithm. In our experiments, we adopt GRPO for policy optimization.
% Recent work has further shown that these two stages serve complementary roles: SFT internalizes structured knowledge and formats, while RL promotes generalization and exploration beyond the supervised distribution~\citep{chu2025sft, zhang2026good}. This SFT-then-RL recipe has been extended beyond instruction following to reasoning~\citep{guo2025deepseek}, tool use~\citep{qian2025toolrl}, and agentic decision-making. 
% \ours\ adopts this established paradigm but applies it to a novel target: rather than fine-tuning the execution agent itself, we train a separate context augmentation model $\pi^C_\theta$ via SFT and GRPO to generate task-specific context that steers frozen downstream agents.
% \zx{the last two sentences are disconnected to this paper? Instead, we should add one sentence on the related works' connection to this paper/proposed method}

\section{Preliminaries}
\label{sec:preli}

% In this section, we introduce the foundational formulations and notations used throughout the paper.

\subsection{LLM Reinforcement Learning.}

The application of reinforcement learning to LLMs became popular with reinforcement learning from human feedback (RLHF)~\citep{ouyang2022training, leike2018scalable, askell2021general, bai2022training, rafailov2023direct}. In this framework, an LLM is modeled as a stochastic policy $\pi_\theta(y \mid x)$ that generates tokens autoregressively. Human preference data are first collected to train a reward model, which is then used to optimize the policy via reinforcement learning.
% , typically with proximal policy optimization (PPO)~\citep{schulman2017proximal}.

% However, training a separate reward model can be computationally expensive and operationally complex. Subsequent work therefore seeks to reduce this overhead by leveraging group-relative advantages~\citep{shao2024deepseekmath} or by treating the policy model itself as an implicit reward signal~\citep{rafailov2023direct}, thereby eliminating the need for an explicit reward model. This paradigm has substantially improved alignment, helpfulness, and safety in instruction-following models.

More recently, RL-based post-training has been extended to enhance reasoning ability, tool use, and long-horizon decision-making in agentic settings. 
These approaches frame text generation as sequential decision-making with downstream task rewards, rather than purely token-level prediction, as introduced in \Cref{subsec:llm_decision}.

\subsection{LLM Agent for Decision Making.}
\label{subsec:llm_decision}
As LLMs grow increasingly capable, deploying them as agents in sequential decision-making settings has become a prominent research direction~\citep{nakano2021webgpt, yao2023react}. We formalize this setting by modeling an LLM agent as a policy $\pi$ within a Partially Observable Markov Decision Process (POMDP), defined by the tuple 
\begin{equation}
\label{eq:pomdp}
    M = (\mathcal{S}, \mathcal{A}, \mathcal{O}, P, R, \gamma),
\end{equation}
where $\mathcal{S}$ is the latent state space, $\mathcal{A}$ is the action space, $\mathcal{O}$ is the observation space, $P(s_{t+1} \mid s_t, a_t)$ is the transition dynamics, $R$ is the reward function, and $\gamma$ is the discount factor.

An initial task description $q$ is sampled from the task distribution $q\sim\mc{D}$, which produces a state $s_0$. Based on $q$ and historical observation, the agent makes an action and receives an observation from the environment. Denote the agent's history as 
$$h_0 = q, \quad  h_t = (q, a_0, o_1, a_1, \dots, o_{t-1}, a_{t-1}, o_t).$$
At each time step $t$, the task execution LLM agent $\pi_{\theta}^E(a_t \mid h_t)$ consumes the full history $h_t$ and produces the next action $a_t \in \mathcal{A}$. The environment then transitions to a new state $s_{t+1}$ according to $P$ and returns the subsequent observation $o_{t+1}\in\mc{O}$.

This process continues until the interaction terminates after $T$ steps, yielding a complete episode trajectory 
\begin{equation}
\label{eq:tau}
    \tau = (q, a_0, o_1, a_1, \dots, o_{T-1}, a_{T-1}, o_T).
\end{equation}
% \linbo{preview how to add $c$ to have a better $q$}
Then a scalar reward $r = R(\tau)$ is assigned to the entire trajectory, reflecting the overall quality of the agent's behavior over the episode. In this work, we propose a context augmentation method to add auxiliary context $c$ into $q$, so that the expected reward can be improved on $\mc{D}$.

\section{Our Proposed Method}
\label{sec:clear}

\begin{figure*}
    \centering
    \includegraphics[width=2\columnwidth]{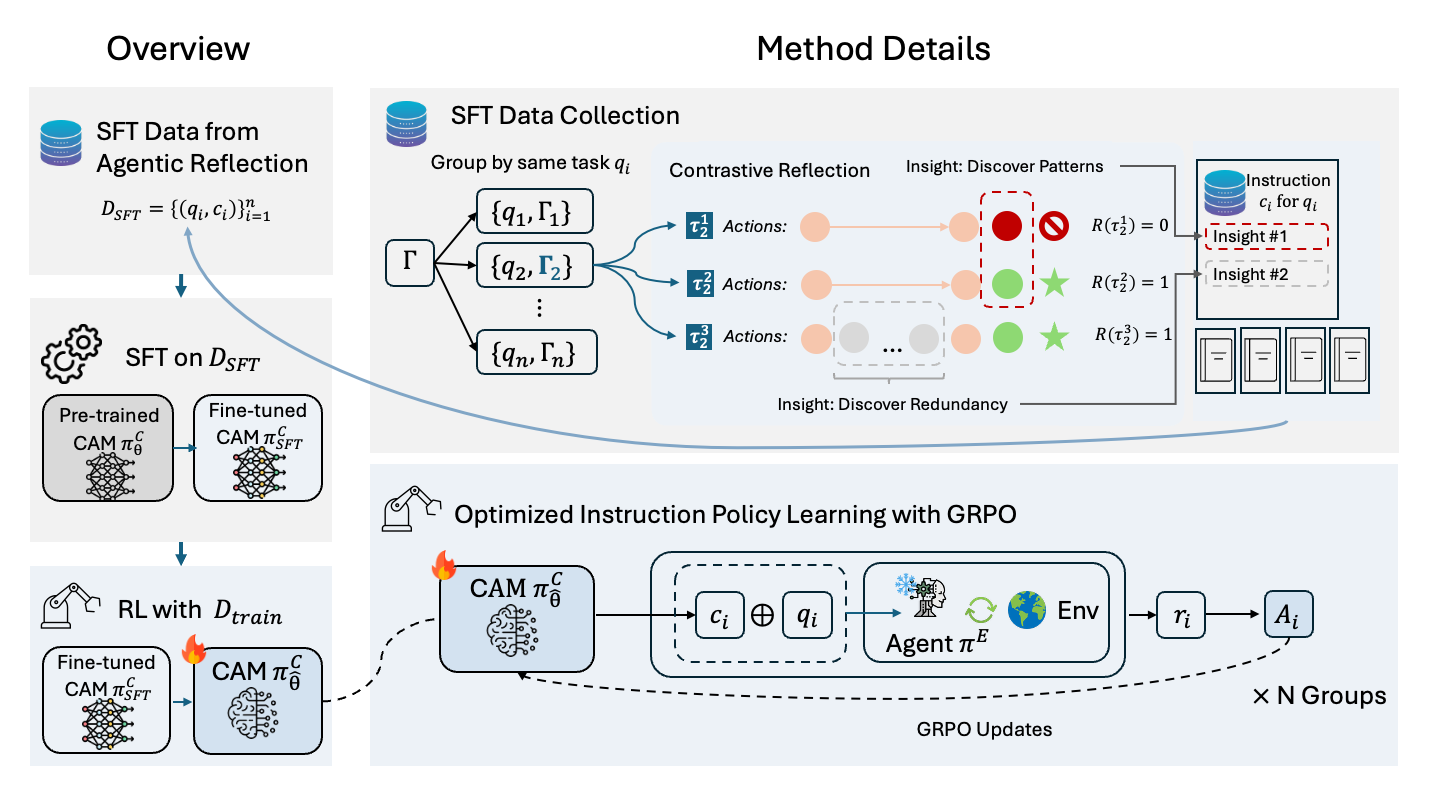}
    \caption{\ours\ training framework design. First, we execute each task $q_i\sim\mc{D}_{\text{train}}$ for $m$ times and collect groups of replay $\Gamma_i$ for $q_i$. We employ reflection agents $\pi^R$ to generate instance-level instruction $c_i$ for each $q_i$, collected into $\mc{D}_{\text{SFT}}$. Then, we initialize CAM from an open-source LLM and perform SFT on $\mc{D}_{\text{SFT}}$. Finally, we further perform RL on the trained CAM, which leverages the reward signal from $\pi^E$ for policy update of CAM.}
    \label{fig:clear-workflow}
\end{figure*}

Many practical LLM-based agents (e.g., \cite{yang2024swe, xia2025live, liu2025migrationbench}) are built on top of proprietary foundation models such as OpenAI GPT models~\citep{singh2025openai}, Anthropic Claude~\citep{claude_4_model_card, claude_4_5_model_card}, and Google Gemini~\citep{team2023gemini}. 
Although these agentic systems are often deployed through open-source agent frameworks such as Strands Agents\footnoteurl{https://strandsagents.com/latest/}, LangGraph\footnoteurl{https://www.langchain.com/langgraph}, and OpenHands~\citep{wang2025openhandssoftwareagentsdk}, the underlying foundation models remain closed-source. 
As a result, their internal parameters are inaccessible, limiting the feasibility of weight-level adaptation.

In this work, we propose a unified context augmentation framework that operates without modifying the LLM agents' weights. Our approach is compatible with both proprietary models and open-source models such as Qwen~\citep{yang2025qwen3}, DeepSeek~\citep{liu2025deepseek}, Olmo~\citep{olmo2025olmo}, and Kimi~\citep{team2025kimi}. Instead of updating model parameters, we improve agent performance by augmenting the context via contrastive learning from past experience. When a task execution agent $\pi^E(\cdot)$ performs a task $q$, we augment its task description $q$ with additional context $c$ produced by CAM. Formally, we define a \emph{replay buffer} 
\[
\Gamma = \big\{(\tau_1, R(\tau_1)), \dots, (\tau_n, R(\tau_n))\big\}
\]
as a collection of past trajectories and their corresponding outcome rewards, where $\tau_i$ is defined in \Cref{eq:tau}. Let $q \sim \mathcal{D}$ be an initial task description sampled from task distribution $\mathcal{D}$. We define a \emph{context augmentation model} that maps $q$ to an auxiliary context $c$ that is appended to $q$ before action generation from $\pi^E$. In other words, execution agent $\pi^E$ will have $q\oplus c$ as its task description, where $\oplus$ denote concatenation. 

We define the CAM as $\pi^C_\theta(\cdot)$ parameterized by $\theta$. 
Given a task $q$, the model generates additional context 
$
c = \pi^C_\theta(q).
$
Our objective is to learn an optimal $\pi^C_\theta(\cdot)$ from $\Gamma$ such that the expected return of the task execution agent $\pi^{E}$ is maximized on $\mathcal{D}_{\text{train}}$.

To achieve this objective, we propose \ours\ (\textbf{C}ontrastive \textbf{L}earning of \textbf{E}xperience via \textbf{A}gentic \textbf{R}eflection), a three-phase training framework that combines contrastive learning, agentic reflection, SFT, and RL to optimize the CAM $\pi^C_\theta$.  In Phase 0, we employ a reflection agent to perform contrastive analysis over past execution trajectories and generate training data for SFT. 
In Phase 1, we fine-tune an open-source LLM using SFT as a warm-up stage. 
In Phase 2, we further optimize the model via RL to directly maximize the expected return of the task execution agent:
% \begin{equation}
% \label{eq:rl}
%     \max_\theta J({\theta})
% =
% \max_\theta\mathbb{E}_{\substack{q \sim \mathcal{D} \\
% c \sim \pi_\theta^C(q) \\
% \tau \sim \pi^E(\cdot \mid q \oplus c)}}
% \big[ R(\tau) \big].
% \end{equation}

\begin{align}
\label{eq:rl}
    {}&\max_\theta J({\theta})\notag\\
=&
\max_\theta\mathbb{E}_{{q \sim \mathcal{D}_{\text{train}},\,
c \sim \pi_\theta^C(q),\,
\tau \sim \pi^E(\cdot \mid q \oplus c)}}
\big[ R(\tau) \big].
\end{align}

Intuitively, this objective encourages the CAM to generate useful context that improves the execution agent’s expected performance. A concurrent work~\citep{asawa2025train} designs a similar RL pipeline to train an advisor model, but doesn't perform contrastive learning using agentic reflection and SFT. However, as discussed in \Cref{app:ablation}, all three phases in \ours\ can bring non-trivial performance improvement, making it a more comprehensive framework for agent refinement.
We now introduce \ours\ in details.

\subsection{Agentic Reflection via Contrastive Learning} 
% \linbo{write a paragraph to emphasize that this is different from prompt optimization.}

% In this section, we introduce how to generate good training data for SFT using contrastive learning. 

% Learning solely from positive or negative trajectories is insufficient for robust agent refinement. 
% If the model only observes successful trajectories, it cannot disentangle which behaviors are causally responsible for success versus which are incidental. 
% Conversely, if it only observes failed trajectories, it lacks a reference signal to identify what was missing or incorrect. 
% Therefore, effective refinement requires \emph{contrastive signals} that highlight the behavioral differences between successful and failed executions.

Learning from a single trajectory is insufficient for robust agent refinement. A single execution provides only a narrow and potentially noisy view of the decision process.
% , making it difficult to determine which behaviors are causally important versus incidental artifacts of that particular run. 
Therefore, refinement should leverage \emph{multiple trajectories} for the same task. See \Cref{app:ablation} for an ablation study.
% The above observation is supported by the ablation study in \Cref{app:ablation}.
% When both successful and unsuccessful trajectories are available, they provide natural \emph{contrastive signals} that highlight the behavioral differences leading to success or failure. When only one type of trajectory is available, useful signals can still be extracted by summarizing the key steps associated with success or identifying the critical mistakes that lead to failure. 

To achieve this, we introduce a {reflection agent} $\pi^R$ that performs contrastive analysis over the {replay buffer} $\Gamma$ for data generation. Its objective is to extract high-value insights that explain the behavioral distinctions among multiple trajectories. To enable scalable analysis, the reflection agent is equipped with a shell tool that allows it to selectively read trajectory files. This design is particularly important when trajectories are large and cannot be loaded entirely into context. We provide the prompts for $\pi^R$ in \Cref{app:prompt}.

To fully leverage the benefits of contrastive analysis, we execute each task multiple times to obtain a set of trajectories corresponding to the same task instance. These trajectories capture diverse execution behaviors and outcomes, providing a rich resource for identifying task-specific decision patterns through contrastive comparison.

% Recall that a \emph{replay buffer}
% $
% \Gamma = \{(\tau_i, R(\tau_i))\}_{i=1}^n
% $
% consists of past trajectories $\tau_i$ and their corresponding outcome rewards $R(\tau_i)$. 

% To better leverage contrastive learning, we group trajectories in $\Gamma$ based on their task $q$. 
Specifically, for each task instance $q_i \sim \mathcal{D}_{\text{train}}$, we execute the task $m$ times to obtain trajectories $\tau_i^1, \dots, \tau_i^m$ and their corresponding rewards $r_i^1, \dots, r_i^m$. 
These trajectories are then organized into a grouped replay buffer
$
\Gamma_i = \{(\tau_i^{1}, r_i^{1}), \dots, (\tau_i^{m}, r_i^{m})\}.
$
We obtain $
c_i = \pi^R(\Gamma_i, q_{i})
$
by applying a reflection agent $\pi^R$ to analyze the replay buffer $\Gamma_i$ and summarize helpful context.
Intuitively speaking, $c_i$ can be viewed as an additional instruction to complete $q_i$.
 Up to this point, the generated pairs $(q_i, c_i)$ form a high-quality SFT dataset, which will be used to train the augmentation model $\pi_\theta^C$.

\subsection{Training Framework}
\label{subsec:training}
In this subsection, we adopt the standard post-training paradigm widely used in LLM alignment~\citep{ouyang2022training, guo2025deepseek, liu2025deepseek, yang2025qwen3}: a two-stage framework consisting of SFT followed by RL.

\paragraph{SFT.}

Using the data collection pipeline described previously, we obtain a supervised dataset 
$
\mathcal{D}_{\text{SFT}} = \{(q_i, c_i)\}_i.
$
We initialize the context augmentation model $\pi^C_\theta$ with a pre-trained LLM parameterized by $\theta$. 
We then fine-tune the model on the supervised dataset $\mathcal{D}_{\text{SFT}}$ to obtain updated model $\pi^C_{\text{SFT}}$. 
The resulting model will be served as the initialization for the subsequent RL stage.

\paragraph{RL.}
In this phase, we further optimize $\pi^C_{\text{SFT}}$ using reinforcement learning to directly maximize expected task reward.
The training objective is introduced in \Cref{eq:rl}. Note that (i) In \Cref{eq:rl}, parameters $\theta$ in  $\pi^C_\theta$ are the only trainable parameters, and the execution agent $\pi^E$ will be frozen. (ii) The reward signal for $\pi^C_{\theta}$ is the same as the reward for running the execution agent $\pi^E$ with $q\oplus c$, where $c\sim\pi^C_{\theta}(q)$. We optimize $\pi^C_{\theta}$ using policy gradient methods. Specifically in our experiments, we adopt GRPO as the policy optimization algorithm. See \Cref{fig:clear-workflow} for an illustration of the workflow. 
% In \Cref{app:ablation}, we show that all components (contrastive learning, SFT, RL) in \ours\ are necessary for best performance.

% Recall that $\pi^{E}(\cdot)$ denotes the task execution agent interacting with the environment $M$ defined in \Cref{eq:pomdp}. 
% Conditioned on the augmented observation $o_0 \oplus c$, the execution agent produces a trajectory $\tau \sim \pi(\cdot \mid o_0 \oplus c)$ and receives cumulative reward $R(\tau)$. 
% We define the training objective of the augmentation policy as maximizing the expected downstream reward:
% \[
% J(\hat{\theta})
% =
% \mathbb{E}_{\substack{o_0 \sim \mathcal{D} \\
% y \sim I_{s,\hat{\theta}}(\cdot \mid o_0) \\
% \tau \sim \pi(\cdot \mid o_0 \oplus y)}}
% \big[ R(\tau) \big].
% \]

% Given a sampled initial observation $o_0 \sim \mathcal{D}_{\text{train}}$, the augmentation model generates a single-turn output
% \[
% c = \pi^C_{\text{SFT}}(o_0),
% \]
% which serves as the task-specific instruction appended to $o_0$.

% This two-stage training paradigm combines the stability of supervised learning with the objective-driven optimization of reinforcement learning, resulting in a more robust and performance-driven augmentation model. 

\subsection{Comparison to Existing Work}
% \subsubsection{Comparison to ACE}
% \red{@Guande \& Yawei: Need review}
MIPRO~\citep{opsahl2024optimizing}, GEPA~\citep{agrawal2025gepa} and  ACE~\citep{zhang2025agenticcontextengineeringevolving} 
are representative approaches for automatic prompt optimization. We refer to these methods as \emph{the compared methods} throughout this section.
% is a related work that expands the agent context using a learned playbook generated by a reflector and a curator. 
While these approaches have demonstrated promising results,
our \ours\ framework differs from \emph{the compared methods} in several key aspects.

{
% \color{red}
\textbf{First}, all \emph{the compared methods} are training-free frameworks that relies entirely on off-the-shelf LLMs acting as the reflector and curator to generate the playbook and instructions. 
In contrast, \ours\ performs parametric learning: we train a context augmentation model $\pi^C_\theta$ using SFT followed by RL.

\textbf{Second}, the reflection agent $\pi^R$ used in Phase~0 of \ours\ is inspired by \emph{the compared methods} but differs substantially in design. 
Reflectors in \emph{the compared methods} are implemented as a single LLM call, whereas our $\pi^R$ is an agentic system equipped with tools for trajectory inspection and analysis. 
Moreover, our $\pi^R$ explicitly performs contrastive reasoning over multiple trajectories to extract useful instructions, which is not a focus of \emph{the compared methods}.

\textbf{Third}, the prompt templates used by \emph{the compared methods} for the reflector and curator are benchmark-specific. 
For example, ACE's prompts include explicit instructions tailored to the AppWorld tasks and are not designed to generalize across benchmarks\footnoteurl{https://github.com/ace-agent/ace-appworld/tree/main/experiments/prompts}. 
In contrast, the prompt template used by our reflection agent is general and benchmark-agnostic (see \Cref{app:prompt}).
Despite ACE employing benchmark-specific prompt engineering for AppWorld, \ours\ consistently outperforms ACE as shown in \Cref{tab:aw}.
}

We also discuss the comparison to RAG in \Cref{app:rag}.

% \subsubsection{Comparison to \textsc{GenRead}}
% \textsc{GenRead}~\cite{yu2022generate} proposed an approach of \emph{generate-then-read}, instead of the traditional approach of \emph{retreive-then-read}. \textsc{GenRead} first prompts a LLM to generate related context based on a given question, and then
% reads the generated context to produce the final answer.
\section{Experiments}
To evaluate our \ours\ framework, we conduct experiments on the AppWorld~\citep{trivedi2024appworld} and WebShop~\citep{yao2022webshop} dataset.

\subsection{Experiment Setting}
We introduce our experiment setting for agentic data collection phase as follows and leave the training details of SFT and RL to \Cref{app:training}.

% \subsubsection{Agentic Data Collection}
\begin{table*}[t]
\centering
\begin{tabular}{l cc cc}
\toprule
% \multirow{2}{*}{Model for $\pi^E$} & 
\multirow{2}{*}{Method} 
& \multicolumn{2}{c}{TGC $\uparrow$} 
& \multicolumn{2}{c}{SGC $\uparrow$} \\
\cmidrule(lr){2-3} \cmidrule(lr){4-5}
&  Avg & Pass@3 & Avg & Pass@3 \\
\midrule

% \multirow{6}{*}{\cmd{Claude-Sonnet-4}}
Baseline & 72.62(2.59) & 86.90 & 52.38(2.73) & 66.07  \\
RAG & 72.02(2.15) & 86.31 & 54.67(3.72) & 71.43  \\
% & \textsc{GenRead} & 64.60(1.92) & 73.21 & 44.84(3.28) & 57.14\\
ACE~\cite{zhang2025agenticcontextengineeringevolving} 
& 74.40(3.57) & 85.71 & 58.93(6.19) & 73.21 \\
GEPA~\cite{agrawal2025gepa} & 75.99(0.74) & 89.88 & 57.14(3.86) & 78.57\\
MIPRO~\cite{opsahl2024optimizing} & 74.40(1.46) & 88.10 & 55.36(1.46) & 80.36\\
\ours~(ours) & \textbf{81.15}(2.48) & \textbf{91.67} & \textbf{66.67}(4.49) & \textbf{82.14} \\

% \cmidrule(l){1-6}

% \multirow{4}{*}{\cmd{DeepSeek-V3.1}}
% & Baseline &  41.27(3.05) & 64.29 & 19.05(2.06) & 26.79  \\
% & RAG & 33.33(1.57) & 53.57 & 10.71(4.72) & 21.43  \\
% % & \textsc{GenRead} & 26.98(2.68) & 42.26 & 13.69(2.73) & 23.21\\
% & ACE
% % ~\cite{zhang2025agenticcontextengineeringevolving} 
% & 32.54(1.50) & 53.57 & 13.69(1.03) & 28.57\\
% & \ours~(ours) & \textbf{42.95}(2.85) & \textbf{66.07} & \textbf{24.40}(4.49) & \textbf{33.93}  \\

\bottomrule
\end{tabular}
\caption{\textbf{AppWorld} experiments results. Task Goal Completion (TGC) and Scenario Goal Completion (SGC) on the \textsc{Test-N} split are reported. Results are averaged over three runs (standard deviation in parentheses) except the Pass@3 metric.}
\label{tab:aw}

\end{table*}

\paragraph{Execution Agent.}
We adopt the {Strands Agents} framework as the backbone of our agentic system. 
Strands Agents is a lightweight yet powerful SDK for building and deploying AI agents using a model-driven design paradigm. 
It supports a broad range of applications, from simple conversational assistants to complex autonomous workflows, and scales seamlessly from local development to production environments. 
We use \cmd{Claude-Sonnet-4}~\citep{claude_4_model_card} and \cmd{DeepSeek-V3.1}~\citep{liu2024deepseek}, accessed via Amazon Bedrock\footnoteurl{https://aws.amazon.com/bedrock/}, as the foundation models for the execution agent $\pi^E$. 
% We enable the model's thinking mode and set the thinking budget to 2{,}048 tokens. 
% All other hyperparameters are set to their default values. 
We leverage the above agentic framework as the execution agent to run the training dataset of AppWorld and Webshop. To accelerate agent execution, we deploy the agent to Amazon Bedrock AgentCore\footnoteurl{https://aws.amazon.com/bedrock/agentcore/} Runtime, which bootstraps multiple containers in parallel to support high-concurrency rollout execution. We set $m=6$, i.e. for each task in the training set, we run the agent 6 times to collect trajectories. 
We then use their official evaluation harness to compute the outcome reward for each trajectory.

\paragraph{Reflection Agent.}
We use the Strands Agents framework together with \cmd{Claude-Sonnet-4} to build a reflection agent for contrastive analysis. The full prompt used for the reflection agent is provided in \Cref{app:prompt}. Furthermore, we leverage a combinatorial data augmentation technique to enlarge SFT dataset size if insufficient, as detailed in \Cref{app:training}.

\paragraph{CAM.} The CAM $\pi^C_\theta(\cdot)$ is initialized from a \cmd{Qwen/Qwen3-32B} model~\citep{yang2025qwen3}, downloaded from HuggingFace. SFT and RL details can be found in \Cref{app:training}.

\subsection{Baseline and Compared Methods}
\paragraph{Baseline.}
% We consider two baseline approaches for comparison. 
We compare \ours~with the untuned baseline using the execution agent $\pi^{E}_\theta(\cdot)$ and the initial task description $q$ without any context augmentation.

\paragraph{RAG.} To provide a stronger comparison, we also construct a RAG method. 
Specifically, we store all $(q_i, c_i)$ pairs from $\mathcal{D}_{\text{SFT}}$ in a vector database, whose embedding is generated by \cmd{BAAI/bge-base-en-v1.5}~\citep{bge_embedding} from HuggingFace. 
During execution for a new task $q_{\text{new}} \sim \mathcal{D}_{\text{test}}$, we find the most similar task $q_j$ from the training set and retrieve the index according to
$
j = \argmax_{1 \le i \le |\mathcal{D}_{\text{SFT}}|}
\text{Sim}(E(q_i), E(q_{\text{new}})),
$
where $E$ denotes sentence embedding and $\text{Sim}(\cdot,\cdot)$ denotes the cosine similarity.
The corresponding instruction $c_j$ is then appended to the new task $q_{\text{new}}$. 
The execution agent subsequently operates on the augmented context $q_{\text{new}} \oplus c_j$. 
We refer to this approach as the \textsc{RAG}.

\paragraph{The compared methods.} We also report results for ACE~\citep{zhang2025agenticcontextengineeringevolving} on AppWorld dataset. ACE models the context as an evolving playbook that accumulates and refines task-solving strategies through generation, reflection, and curation.
To ensure a fair comparison, we adapt the official ACE GitHub repository\footnoteurl{https://github.com/ace-agent/ace} to the Strands Agents framework and use the same LLM, \cmd{Claude-Sonnet-4}, as in \ours.  
For GEPA and MIPROv2, we use the official implementations by \cmd{dspy} \footnote{https://github.com/stanfordnlp/dspy}. 
% {\color{red}
% \paragraph{GEPA} Introduce how we produce the result for GEPA

% \paragraph{MIPROv2} Introduce how we produce the result for MIPROv2
% }

\paragraph{\ours.}
Given a new task $q_{\text{new}}$, 
% instead of retrieving an instruction from the knowledge base, 
we generate auxiliary context $c$ using our CAM served via vLLM~\citep{kwon2023efficient}:
$
c \sim \pi^C_\theta(q_{\text{new}}).
$
The execution agent $\pi^{{E}}_\theta$ then operates on the augmented description $q_{\text{new}} \oplus c$ to generate a trajectory and receive a reward.

\subsection{Results on AppWorld}
We report experiment results on AppWorld in this subsection. We use the \textsc{Train} split as the training set $\mc{D}_{\text{train}}$ and the \textsc{Test-N} split as the evaluation set $\mc{D}_{\text{test}}$.
These two splits are disjoint and follow the official dataset partition defined in the original paper, ensuring that no data leakage occurs during evaluation. For the execution agent, we use the original system prompt, which is available in the official AppWorld repository\footnoteurl{https://github.com/StonyBrookNLP/appworld/blob/main/experiments/prompts/react_code_agent/_legacy_instructions.txt}. 

% \paragraph{\textsc{GenRead}}\red{write a paragraph about GenRead.}

\paragraph{Metrics.}
We use Task Goal Completion (TGC) and Scenario Goal Completion (SGC) rates from AppWorld \citep{trivedi2024appworld} as our evaluation metrics. 
TGC is defined as the percentage of tasks for which the agent passes all evaluation tests provided by the AppWorld benchmark. 
SGC measures the percentage of task scenarios for which the agent passes all evaluation tests across every task within the scenario. 
We report TGC, SGC (averaged over 3 independent runs), and their pass@3 rates in \Cref{tab:aw}.

\subsection{Results on WebShop-40k}
% \red{@Han: Introduce WebShop-40k. 1. Motivation to curate a subset. 2. How did we curate 40k. 3. How to compute reward. 4. What is the system prompt used.}

As introduced earlier, we leverage Amazon Bedrock AgentCore for scalable rollout collection. This setup, however, imposes a constraint on Docker image size that is incompatible with the original WebShop benchmark: the full WebShop environment includes a local search index spanning millions of product items, which exceeds the image size limit enforced by AgentCore Runtime. To address this, we randomly sampled 40,000 items from the original product pool to construct a lightweight search index, forming the WebShop-40k variant. We then filtered the task set to retain only those tasks whose ground-truth target items exist within this 40k subset, ensuring reward calculation remains identical to the original benchmark formulation. 
We note that WebShop-40k may be inherently easier than the original benchmark, as the reduced search space lowers the difficulty of product retrieval. 
For the baseline agent, we curated a system prompt that describes the task objective, the available tools, and their corresponding descriptions.

Since ACE does not provide benchmark-specific prompts for WebShop dataset, we do not report ACE results on WebShop-40k. 
Other experimental settings are the same as AppWorld evaluation. 
The results are reported in \Cref{tab:webshop}.

\begin{table}[t]
\centering
\begin{tabular}{l cc}
\toprule
Method & Avg. Reward & Std.\\
\midrule
% \multirow{3}{*}{\cmd{Claude-Sonnet-4}}
Baseline & 0.6799 & 0.0119  \\
RAG &  0.7252 & 0.0076\\
% & \textsc{GenRead} & 0.6016(0.0065) \\
\ours~(ours) & \textbf{0.7406} & 0.0044\\
\bottomrule
\end{tabular}
\caption{\textbf{WebShop-40k} experiments results. Averaged reward on the test dataset is reported. Results are averaged over three runs with standard deviation also reported.}
\label{tab:webshop}
\end{table}

\subsection{Discussion}

As shown in \Cref{tab:aw}, \ours\ consistently outperforms all baselines across all models and all metrics, without using any benchmark-specific prompts in data generation and training pipeline. Especially compared to ACE, \ours\ achieves notable gains of $+6.75$ and $+7.74$ in TGC and SGC respectively, despite ACE using AppWorld-specific prompt for their reflector and curator. Similar improvements over the baselines are also observed on WebShop-40k dataset as shown in \Cref{tab:webshop}.

% \paragraph{Ablation Study.}
\textbf{Ablation Study.}
To demonstrate all components in \ours\ are necessary, we perform ablation study in \Cref{app:ablation}.
\Cref{tab:ablation} in \Cref{app:ablation} 
shows that contrastive learning, SFT and RL each brings non-trivial performance improvement. See \Cref{app:ablation} for more details.

\textbf{Latency Study.}
We present a latency study of CAM in \Cref{app:latency}. As shown in \Cref{tab:latency}, the additional overhead introduced by CAM is modest compared to the performance gains.

\textbf{CAM Transferability.}
To study CAM transferability, we conducted additional study using \cmd{DeepSeek-V3.1} as $\pi^E$ in \Cref{tab:deepseek} of \Cref{app:transferability}, while the entire CAM training data is generated from Claude model. We observe that CAM can still consistently outperform all baselines, despite the training and inference mismatch. See \Cref{app:transferability} for details.

\section{Conclusion}

LLM agents are increasingly used in sequential decision-making to complete complex tasks. 
In this paper, we propose \ours, a novel framework that trains a context augmentation model (CAM) to improve agent performance by generating task-relevant context and appending it to the prompt of the execution LLM agent. 
\ours\ first employs a reflection agent to perform contrastive analysis over past execution trajectories and summarize useful context for each observed task. 
These summaries are then used to train the CAM. 
Although \ours\ requires training a smaller CAM, it does not modify the parameters of the execution LLM agent. 
As a result, \ours\ can be applied to a wide range of LLM agent systems regardless of whether the underlying models are open-source or proprietary. 
Extensive experiments show that \ours\ consistently outperforms several strong baselines across multiple benchmarks.

\section*{{Limitations}}
While \ours\ demonstrates strong empirical performance, several limitations remain. 

First, the effectiveness of \ours\ relies on the quality and diversity of the collected experience replay buffer.
If the replays are noisy, sparse, or lack sufficient successful examples, the reflection agent may fail to extract high-quality instructions, which in turn limits the performance of the learned context augmentation model.

Second, although \ours\ avoids modifying the execution agent, it introduces a non-trivial training pipeline involving reflection, SFT and RL, which can be computationally expensive and time-consuming. However, this cost is incurred only once during training. At inference time, \ours\ introduces minimal additional latency, making it practical for deployment in real-world applications. See \cref{app:latency} for a study on inference latency.

% Bibliography entries for the entire Anthology, followed by custom entries
%\bibliography{anthology,custom}
% Custom bibliography entries only
\bibliography{reference}

@article{singh2025openai,
  title={Openai gpt-5 system card},
  author={Singh, Aaditya and Fry, Adam and Perelman, Adam and Tart, Adam and Ganesh, Adi and El-Kishky, Ahmed and McLaughlin, Aidan and Low, Aiden and Ostrow, AJ and Ananthram, Akhila and others},
  journal={arXiv preprint arXiv:2601.03267},
  year={2025}
}

@misc{claude_4_model_card,
  title        = {System Card: Claude Opus 4 \& Claude Sonnet 4},
  author       = {Anthropic},
  year         = {2025},
  url          = {https://www-cdn.anthropic.com/6d8a8055020700718b0c49369f60816ba2a7c285.pdf
},
  note         = {Accessed: 2026-02-02}
}

@misc{claude_4_5_model_card,
  title        = {System Card: Claude Sonnet 4.5},
  author       = {Anthropic},
  year         = {2025},
  url          = {https://www-cdn.anthropic.com/963373e433e489a87a10c823c52a0a013e9172dd.pdf},
  note         = {Accessed: 2026-02-02}
}

@article{team2023gemini,
  title={Gemini: a family of highly capable multimodal models},
  author={Team, Gemini and Anil, Rohan and Borgeaud, Sebastian and Alayrac, Jean-Baptiste and Yu, Jiahui and Soricut, Radu and Schalkwyk, Johan and Dai, Andrew M and Hauth, Anja and Millican, Katie and others},
  journal={arXiv preprint arXiv:2312.11805},
  year={2023}
}

@article{yang2024swe,
  title={Swe-agent: Agent-computer interfaces enable automated software engineering},
  author={Yang, John and Jimenez, Carlos E and Wettig, Alexander and Lieret, Kilian and Yao, Shunyu and Narasimhan, Karthik and Press, Ofir},
  journal={Advances in Neural Information Processing Systems},
  volume={37},
  pages={50528--50652},
  year={2024}
}

@article{xia2025live,
  title={Live-SWE-agent: Can Software Engineering Agents Self-Evolve on the Fly?},
  author={Xia, Chunqiu Steven and Wang, Zhe and Yang, Yan and Wei, Yuxiang and Zhang, Lingming},
  journal={arXiv preprint arXiv:2511.13646},
  year={2025}
}

@article{liu2025migrationbench,
  title={MigrationBench: Repository-Level Code Migration Benchmark from Java 8},
  author={Liu, Linbo and Liu, Xinle and Zhou, Qiang and Chen, Lin and Liu, Yihan and Nguyen, Hoan and Omidvar-Tehrani, Behrooz and Shen, Xi and Huan, Jun and Tripp, Omer and others},
  journal={arXiv preprint arXiv:2505.09569},
  year={2025}
}

@article{nakano2021webgpt,
  title={Webgpt: Browser-assisted question-answering with human feedback},
  author={Nakano, Reiichiro and Hilton, Jacob and Balaji, Suchir and Wu, Jeff and Ouyang, Long and Kim, Christina and Hesse, Christopher and Jain, Shantanu and Kosaraju, Vineet and Saunders, William and others},
  journal={arXiv preprint arXiv:2112.09332},
  year={2021}
}

@article{ouyang2022training,
  title={Training language models to follow instructions with human feedback},
  author={Ouyang, Long and Wu, Jeffrey and Jiang, Xu and Almeida, Diogo and Wainwright, Carroll and Mishkin, Pamela and Zhang, Chong and Agarwal, Sandhini and Slama, Katarina and Ray, Alex and others},
  journal={Advances in neural information processing systems},
  volume={35},
  pages={27730--27744},
  year={2022}
}

@article{yang2025qwen3,
  title={Qwen3 technical report},
  author={Yang, An and Li, Anfeng and Yang, Baosong and Zhang, Beichen and Hui, Binyuan and Zheng, Bo and Yu, Bowen and Gao, Chang and Huang, Chengen and Lv, Chenxu and others},
  journal={arXiv preprint arXiv:2505.09388},
  year={2025}
}

@article{olmo2025olmo,
  title={Olmo 3},
  author={Olmo, Team and Ettinger, Allyson and Bertsch, Amanda and Kuehl, Bailey and Graham, David and Heineman, David and Groeneveld, Dirk and Brahman, Faeze and Timbers, Finbarr and Ivison, Hamish and others},
  journal={arXiv preprint arXiv:2512.13961},
  year={2025}
}

@article{team2025kimi,
  title={Kimi k2: Open agentic intelligence},
  author={Team, Kimi and Bai, Yifan and Bao, Yiping and Chen, Guanduo and Chen, Jiahao and Chen, Ningxin and Chen, Ruijue and Chen, Yanru and Chen, Yuankun and Chen, Yutian and others},
  journal={arXiv preprint arXiv:2507.20534},
  year={2025}
}

@article{liu2025deepseek,
  title={Deepseek-v3. 2: Pushing the frontier of open large language models},
  author={Liu, Aixin and Mei, Aoxue and Lin, Bangcai and Xue, Bing and Wang, Bingxuan and Xu, Bingzheng and Wu, Bochao and Zhang, Bowei and Lin, Chaofan and Dong, Chen and others},
  journal={arXiv preprint arXiv:2512.02556},
  year={2025}
}

@article{guo2025deepseek,
  title={Deepseek-r1: Incentivizing reasoning capability in llms via reinforcement learning},
  author={Guo, Daya and Yang, Dejian and Zhang, Haowei and Song, Junxiao and Wang, Peiyi and Zhu, Qihao and Xu, Runxin and Zhang, Ruoyu and Ma, Shirong and Bi, Xiao and others},
  journal={arXiv preprint arXiv:2501.12948},
  year={2025}
}

@article{shao2024deepseekmath,
  title={Deepseekmath: Pushing the limits of mathematical reasoning in open language models},
  author={Shao, Zhihong and Wang, Peiyi and Zhu, Qihao and Xu, Runxin and Song, Junxiao and Bi, Xiao and Zhang, Haowei and Zhang, Mingchuan and Li, YK and Wu, Yang and others},
  journal={arXiv preprint arXiv:2402.03300},
  year={2024}
}

@article{schulman2017proximal,
  title={Proximal policy optimization algorithms},
  author={Schulman, John and Wolski, Filip and Dhariwal, Prafulla and Radford, Alec and Klimov, Oleg},
  journal={arXiv preprint arXiv:1707.06347},
  year={2017}
}

@article{leike2018scalable,
  title={Scalable agent alignment via reward modeling: a research direction},
  author={Leike, Jan and Krueger, David and Everitt, Tom and Martic, Miljan and Maini, Vishal and Legg, Shane},
  journal={arXiv preprint arXiv:1811.07871},
  year={2018}
}

@article{askell2021general,
  title={A general language assistant as a laboratory for alignment},
  author={Askell, Amanda and Bai, Yuntao and Chen, Anna and Drain, Dawn and Ganguli, Deep and Henighan, Tom and Jones, Andy and Joseph, Nicholas and Mann, Ben and DasSarma, Nova and others},
  journal={arXiv preprint arXiv:2112.00861},
  year={2021}
}

@article{bai2022training,
  title={Training a helpful and harmless assistant with reinforcement learning from human feedback},
  author={Bai, Yuntao and Jones, Andy and Ndousse, Kamal and Askell, Amanda and Chen, Anna and DasSarma, Nova and Drain, Dawn and Fort, Stanislav and Ganguli, Deep and Henighan, Tom and others},
  journal={arXiv preprint arXiv:2204.05862},
  year={2022}
}

@article{rafailov2023direct,
  title={Direct preference optimization: Your language model is secretly a reward model},
  author={Rafailov, Rafael and Sharma, Archit and Mitchell, Eric and Manning, Christopher D and Ermon, Stefano and Finn, Chelsea},
  journal={Advances in neural information processing systems},
  volume={36},
  pages={53728--53741},
  year={2023}
}

@inproceedings{trivedi2024appworld,
  title={Appworld: A controllable world of apps and people for benchmarking interactive coding agents},
  author={Trivedi, Harsh and Khot, Tushar and Hartmann, Mareike and Manku, Ruskin and Dong, Vinty and Li, Edward and Gupta, Shashank and Sabharwal, Ashish and Balasubramanian, Niranjan},
  booktitle={Proceedings of the 62nd Annual Meeting of the Association for Computational Linguistics (Volume 1: Long Papers)},
  pages={16022--16076},
  year={2024}
}

@inproceedings{zheng2024llamafactory,
  title={LlamaFactory: Unified Efficient Fine-Tuning of 100+ Language Models},
  author={Yaowei Zheng and Richong Zhang and Junhao Zhang and Yanhan Ye and Zheyan Luo and Zhangchi Feng and Yongqiang Ma},
  booktitle={Proceedings of the 62nd Annual Meeting of the Association for Computational Linguistics (Volume 3: System Demonstrations)},
  address={Bangkok, Thailand},
  publisher={Association for Computational Linguistics},
  year={2024},
  url={http://arxiv.org/abs/2403.13372}
}

@article{sheng2024hybridflow,
  title   = {HybridFlow: A Flexible and Efficient RLHF Framework},
  author  = {Guangming Sheng and Chi Zhang and Zilingfeng Ye and Xibin Wu and Wang Zhang and Ru Zhang and Yanghua Peng and Haibin Lin and Chuan Wu},
  year    = {2024},
  journal = {arXiv preprint arXiv: 2409.19256}
}

@inproceedings{kwon2023efficient,
  title={Efficient Memory Management for Large Language Model Serving with PagedAttention},
  author={Woosuk Kwon and Zhuohan Li and Siyuan Zhuang and Ying Sheng and Lianmin Zheng and Cody Hao Yu and Joseph E. Gonzalez and Hao Zhang and Ion Stoica},
  booktitle={Proceedings of the ACM SIGOPS 29th Symposium on Operating Systems Principles},
  year={2023}
}

@misc{zhang2025agenticcontextengineeringevolving,
      title={Agentic Context Engineering: Evolving Contexts for Self-Improving Language Models}, 
      author={Qizheng Zhang and Changran Hu and Shubhangi Upasani and Boyuan Ma and Fenglu Hong and Vamsidhar Kamanuru and Jay Rainton and Chen Wu and Mengmeng Ji and Hanchen Li and Urmish Thakker and James Zou and Kunle Olukotun},
      year={2025},
      eprint={2510.04618},
      archivePrefix={arXiv},
      primaryClass={cs.LG},
      url={https://arxiv.org/abs/2510.04618}, 
}

@misc{bge_embedding,
      title={C-Pack: Packaged Resources To Advance General Chinese Embedding}, 
      author={Shitao Xiao and Zheng Liu and Peitian Zhang and Niklas Muennighoff},
      year={2023},
      eprint={2309.07597},
      archivePrefix={arXiv},
      primaryClass={cs.CL}
}

@article{yao2022webshop,
  title={Webshop: Towards scalable real-world web interaction with grounded language agents},
  author={Yao, Shunyu and Chen, Howard and Yang, John and Narasimhan, Karthik},
  journal={Advances in Neural Information Processing Systems},
  volume={35},
  pages={20744--20757},
  year={2022}
}

@misc{wang2025openhandssoftwareagentsdk,
      title={The OpenHands Software Agent SDK: A Composable and Extensible Foundation for Production Agents}, 
      author={Xingyao Wang and Simon Rosenberg and Juan Michelini and Calvin Smith and Hoang Tran and Engel Nyst and Rohit Malhotra and Xuhui Zhou and Valerie Chen and Robert Brennan and Graham Neubig},
      year={2025},
      eprint={2511.03690},
      archivePrefix={arXiv},
      primaryClass={cs.SE},
      url={https://arxiv.org/abs/2511.03690}, 
}

@article{liu2024deepseek,
  title={Deepseek-v3 technical report},
  author={Liu, Aixin and Feng, Bei and Xue, Bing and Wang, Bingxuan and Wu, Bochao and Lu, Chengda and Zhao, Chenggang and Deng, Chengqi and Zhang, Chenyu and Ruan, Chong and others},
  journal={arXiv preprint arXiv:2412.19437},
  year={2024}
}

@article{gao2023retrieval,
  title={Retrieval-augmented generation for large language models: A survey},
  author={Gao, Yunfan and Xiong, Yun and Gao, Xinyu and Jia, Kangxiang and Pan, Jinliu and Bi, Yuxi and Dai, Yixin and Sun, Jiawei and Wang, Haofen and Wang, Haofen and others},
  journal={arXiv preprint arXiv:2312.10997},
  volume={2},
  number={1},
  pages={32},
  year={2023}
}

@article{kaplan2020scaling,
  title={Scaling laws for neural language models},
  author={Kaplan, Jared and McCandlish, Sam and Henighan, Tom and Brown, Tom B and Chess, Benjamin and Child, Rewon and Gray, Scott and Radford, Alec and Wu, Jeffrey and Amodei, Dario},
  journal={arXiv preprint arXiv:2001.08361},
  year={2020}
}

@article{wei2022emergent,
  title={Emergent abilities of large language models},
  author={Wei, Jason and Tay, Yi and Bommasani, Rishi and Raffel, Colin and Zoph, Barret and Borgeaud, Sebastian and Yogatama, Dani and Bosma, Maarten and Zhou, Denny and Metzler, Donald and others},
  journal={arXiv preprint arXiv:2206.07682},
  year={2022}
}

@article{xi2025rise,
  title={The rise and potential of large language model based agents: A survey},
  author={Xi, Zhiheng and Chen, Wenxiang and Guo, Xin and He, Wei and Ding, Yiwen and Hong, Boyang and Zhang, Ming and Wang, Junzhe and Jin, Senjie and Zhou, Enyu and others},
  journal={Science China Information Sciences},
  volume={68},
  number={2},
  pages={121101},
  year={2025},
  publisher={Springer}
}

@article{black2024pi_0,
  title={$\pi_0$: A Vision-Language-Action Flow Model for General Robot Control},
  author={Black, Kevin and Brown, Noah and Driess, Danny and Esmail, Adnan and Equi, Michael and Finn, Chelsea and Fusai, Niccolo and Groom, Lachy and Hausman, Karol and Ichter, Brian and others},
  journal={arXiv preprint arXiv:2410.24164},
  year={2024}
}

@article{talebirad2023multi,
  title={Multi-agent collaboration: Harnessing the power of intelligent llm agents},
  author={Talebirad, Yashar and Nadiri, Amirhossein},
  journal={arXiv preprint arXiv:2306.03314},
  year={2023}
}

@inproceedings{wang2024executable,
  title={Executable code actions elicit better llm agents},
  author={Wang, Xingyao and Chen, Yangyi and Yuan, Lifan and Zhang, Yizhe and Li, Yunzhu and Peng, Hao and Ji, Heng},
  booktitle={Forty-first International Conference on Machine Learning},
  year={2024}
}

@article{suzgun2025dynamic,
  title={Dynamic cheatsheet: Test-time learning with adaptive memory},
  author={Suzgun, Mirac and Yuksekgonul, Mert and Bianchi, Federico and Jurafsky, Dan and Zou, James},
  journal={arXiv preprint arXiv:2504.07952},
  year={2025}
}

@article{lewis2020retrieval,
  title={Retrieval-augmented generation for knowledge-intensive nlp tasks},
  author={Lewis, Patrick and Perez, Ethan and Piktus, Aleksandra and Petroni, Fabio and Karpukhin, Vladimir and Goyal, Naman and K{\"u}ttler, Heinrich and Lewis, Mike and Yih, Wen-tau and Rockt{\"a}schel, Tim and others},
  journal={Advances in neural information processing systems},
  volume={33},
  pages={9459--9474},
  year={2020}
}

@inproceedings{ma2023query,
  title={Query rewriting in retrieval-augmented large language models},
  author={Ma, Xinbei and Gong, Yeyun and He, Pengcheng and Zhao, Hai and Duan, Nan},
  booktitle={Proceedings of the 2023 Conference on Empirical Methods in Natural Language Processing},
  pages={5303--5315},
  year={2023}
}

@inproceedings{peng2024large,
  title={Large language model based long-tail query rewriting in taobao search},
  author={Peng, Wenjun and Li, Guiyang and Jiang, Yue and Wang, Zilong and Ou, Dan and Zeng, Xiaoyi and Xu, Derong and Xu, Tong and Chen, Enhong},
  booktitle={Companion Proceedings of the ACM Web Conference 2024},
  pages={20--28},
  year={2024}
}

@inproceedings{huang2025ket,
  title={Ket-rag: A cost-efficient multi-granular indexing framework for graph-rag},
  author={Huang, Yiqian and Zhang, Shiqi and Xiao, Xiaokui},
  booktitle={Proceedings of the 31st ACM SIGKDD Conference on Knowledge Discovery and Data Mining V. 2},
  pages={1003--1012},
  year={2025}
}

@article{yu2022generate,
  title={Generate rather than retrieve: Large language models are strong context generators},
  author={Yu, Wenhao and Iter, Dan and Wang, Shuohang and Xu, Yichong and Ju, Mingxuan and Sanyal, Soumya and Zhu, Chenguang and Zeng, Michael and Jiang, Meng},
  journal={arXiv preprint arXiv:2209.10063},
  year={2022}
}

@inproceedings{shao2023enhancing,
  title={Enhancing retrieval-augmented large language models with iterative retrieval-generation synergy},
  author={Shao, Zhihong and Gong, Yeyun and Shen, Yelong and Huang, Minlie and Duan, Nan and Chen, Weizhu},
  booktitle={Findings of the Association for Computational Linguistics: EMNLP 2023},
  pages={9248--9274},
  year={2023}
}

@article{jin2025search,
  title={Search-r1: Training llms to reason and leverage search engines with reinforcement learning},
  author={Jin, Bowen and Zeng, Hansi and Yue, Zhenrui and Yoon, Jinsung and Arik, Sercan and Wang, Dong and Zamani, Hamed and Han, Jiawei},
  journal={arXiv preprint arXiv:2503.09516},
  year={2025}
}

@article{mei2025survey,
  title={A survey of context engineering for large language models},
  author={Mei, Lingrui and Yao, Jiayu and Ge, Yuyao and Wang, Yiwei and Bi, Baolong and Cai, Yujun and Liu, Jiazhi and Li, Mingyu and Li, Zhong-Zhi and Zhang, Duzhen and others},
  journal={arXiv preprint arXiv:2507.13334},
  year={2025}
}

@article{wang2024survey,
  title={A Survey on Large Language Model based Autonomous Agents},
  author={Wang, Lei and Ma, Chen and Feng, Xueyang and Zhang, Zeyu and Yang, Hao and Zhang, Jingsen and Chen, Zhiyuan and Tang, Jiakai and Chen, Xu and Lin, Yankai and others},
  journal={Frontiers of Computer Science},
  volume={18},
  number={6},
  pages={186345},
  year={2024},
  publisher={Springer}
}

@article{wei2022chain,
  title={Chain-of-thought prompting elicits reasoning in large language models},
  author={Wei, Jason and Wang, Xuezhi and Schuurmans, Dale and Bosma, Maarten and Xia, Fei and Chi, Ed and Le, Quoc V and Zhou, Denny and others},
  journal={Advances in neural information processing systems},
  volume={35},
  pages={24824--24837},
  year={2022}
}

@inproceedings{yao2023react,
  title={ReAct: Synergizing Reasoning and Acting in Language Models},
  author={Yao, Shunyu and Zhao, Jeffrey and Yu, Dian and Du, Nan and Shafran, Izhak and Narasimhan, Karthik and Cao, Yuan},
  booktitle={International Conference on Learning Representations},
  year={2023}
}

@article{yao2023tree,
  title={Tree of thoughts: Deliberate problem solving with large language models},
  author={Yao, Shunyu and Yu, Dian and Zhao, Jeffrey and Shafran, Izhak and Griffiths, Tom and Cao, Yuan and Narasimhan, Karthik},
  journal={Advances in neural information processing systems},
  volume={36},
  pages={11809--11822},
  year={2023}
}

@inproceedings{shinn2023reflexion,
  title={Reflexion: Language Agents with Verbal Reinforcement Learning},
  author={Shinn, Noah and Cassano, Federico and Gopinath, Ashwin and Sheshadri, Karthik and Narasimhan, Karthik and Yao, Shunyu and others},
  booktitle={Advances in Neural Information Processing Systems},
  volume={36},
  year={2023}
}

@article{wang2023voyager,
  title={Voyager: An Open-Ended Embodied Agent with Large Language Models},
  author={Wang, Guanzhi and Xie, Yuqi and Jiang, Yunfan and Mandlekar, Ajay and Xiao, Chaowei and Zhu, Yuke and Fan, Linxi and Anandkumar, Anima},
  journal={arXiv preprint arXiv:2305.16291},
  year={2023}
}

@article{schick2023toolformer,
  title={Toolformer: Language models can teach themselves to use tools},
  author={Schick, Timo and Dwivedi-Yu, Jane and Dess{\`\i}, Roberto and Raileanu, Roberta and Lomeli, Maria and Hambro, Eric and Zettlemoyer, Luke and Cancedda, Nicola and Scialom, Thomas},
  journal={Advances in neural information processing systems},
  volume={36},
  pages={68539--68551},
  year={2023}
}

@article{patil2024gorilla,
  title={Gorilla: Large language model connected with massive apis},
  author={Patil, Shishir G and Zhang, Tianjun and Wang, Xin and Gonzalez, Joseph E},
  journal={Advances in Neural Information Processing Systems},
  volume={37},
  pages={126544--126565},
  year={2024}
}

@inproceedings{li2023camel,
  title={CAMEL: Communicative Agents for ``Mind'' Exploration of Large Language Model Society},
  author={Li, Guohao and Hammoud, Hasan Abed Al Kader and Itani, Hani and Khizbullin, Dmitrii and Ghanem, Bernard},
  booktitle={Advances in Neural Information Processing Systems},
  volume={36},
  year={2023}
}

@inproceedings{hong2024metagpt,
  title={MetaGPT: Meta Programming for A Multi-Agent Collaborative Framework},
  author={Hong, Sirui and Zhuge, Mingchen and Chen, Jonathan and Zheng, Xiawu and Cheng, Yuheng and Zhang, Ceyao and Wang, Jinlin and Wang, Zili and Yau, Steven Ka Shing and Lin, Zijuan and others},
  booktitle={International Conference on Learning Representations},
  year={2024}
}

@inproceedings{qian2024chatdev,
  title={ChatDev: Communicative Agents for Software Development},
  author={Qian, Chen and Liu, Wei and Liu, Hongzhang and Chen, Nuo and Dang, Yufan and Li, Jiahao and Yang, Cheng and Chen, Weize and Su, Yusheng and Cong, Xin and others},
  booktitle={Proceedings of the 62nd Annual Meeting of the Association for Computational Linguistics},
  year={2024}
}

@inproceedings{wu2024autogen,
  title={AutoGen: Enabling Next-Gen LLM Applications via Multi-Agent Conversation},
  author={Wu, Qingyun and Bansal, Gagan and Zhang, Jieyu and Wu, Yiran and Li, Beibin and Zhu, Erkang and Jiang, Li and Zhang, Xiaoyun and Zhang, Shaokun and Liu, Jiale and others},
  booktitle={Conference on Language Modeling},
  year={2024}
}

@misc{anthropic2025mcp,
  title={Introducing the Model Context Protocol},
  author={{Anthropic}},
  year={2025},
  note={\url{https://www.anthropic.com/news/model-context-protocol}}
}

@misc{openai2025operator,
  title={Introducing Operator},
  author={{OpenAI}},
  year={2025},
  note={\url{https://openai.com/index/introducing-operator/}}
}

@article{agrawal2025gepa,
  title={Gepa: Reflective prompt evolution can outperform reinforcement learning},
  author={Agrawal, Lakshya A and Tan, Shangyin and Soylu, Dilara and Ziems, Noah and Khare, Rishi and Opsahl-Ong, Krista and Singhvi, Arnav and Shandilya, Herumb and Ryan, Michael J and Jiang, Meng and others},
  journal={arXiv preprint arXiv:2507.19457},
  year={2025}
}

@article{asawa2025train,
  title={How to train your advisor: Steering black-box llms with advisor models},
  author={Asawa, Parth and Zhu, Alan and O'Neill, Abby and Zaharia, Matei and Dimakis, Alexandros G and Gonzalez, Joseph E},
  journal={arXiv preprint arXiv:2510.02453},
  year={2026}
}

@inproceedings{lingam2025enhancing,
  title={Enhancing language model agents using diversity of thoughts},
  author={Lingam, Vijay and Tehrani, Behrooz Omidvar and Sanghavi, Sujay and Gupta, Gaurav and Ghosh, Sayan and Liu, Linbo and Huan, Jun and Deoras, Anoop},
  booktitle={The Thirteenth International Conference on Learning Representations},
  year={2025}
}

@inproceedings{ahmadian2024back,
  title={Back to basics: Revisiting REINFORCE-style optimization for learning from human feedback in LLMs},
  author={Ahmadian, Arash and Cremer, Chris and Gall{\'e}, Matthias and Fadaee, Marzieh and Kreutzer, Julia and Pietquin, Olivier and {\"U}st{\"u}n, Ahmet and Hooker, Sara},
  booktitle={Proceedings of the 62nd Annual Meeting of the Association for Computational Linguistics (Volume 1: Long Papers)},
  pages={12248--12267},
  year={2024}
}

@inproceedings{zhang2021sample,
  title={Sample efficient reinforcement learning with REINFORCE},
  author={Zhang, Junzi and Kim, Jongho and O'Donoghue, Brendan and Boyd, Stephen},
  booktitle={Proceedings of the AAAI conference on artificial intelligence},
  volume={35},
  pages={10887--10895},
  year={2021}
}

@article{yu2025dapo,
  title={Dapo: An open-source llm reinforcement learning system at scale},
  author={Yu, Qiying and Zhang, Zheng and Zhu, Ruofei and Yuan, Yufeng and Zuo, Xiaochen and Yue, Yu and Dai, Weinan and Fan, Tiantian and Liu, Gaohong and Liu, Lingjun and others},
  journal={arXiv preprint arXiv:2503.14476},
  year={2025}
}

@article{yue2025vapo,
  title={Vapo: Efficient and reliable reinforcement learning for advanced reasoning tasks},
  author={Yue, Yu and Yuan, Yufeng and Yu, Qiying and Zuo, Xiaochen and Zhu, Ruofei and Xu, Wenyuan and Chen, Jiaze and Wang, Chengyi and Fan, TianTian and Du, Zhengyin and others},
  journal={arXiv preprint arXiv:2504.05118},
  year={2025}
}

@article{zheng2025group,
  title={Group sequence policy optimization},
  author={Zheng, Chujie and Liu, Shixuan and Li, Mingze and Chen, Xiong-Hui and Yu, Bowen and Gao, Chang and Dang, Kai and Liu, Yuqiong and Men, Rui and Yang, An and others},
  journal={arXiv preprint arXiv:2507.18071},
  year={2025}
}

@inproceedings{li2024learning,
  title={Learning to rewrite prompts for personalized text generation},
  author={Li, Cheng and Zhang, Mingyang and Mei, Qiaozhu and Kong, Weize and Bendersky, Michael},
  booktitle={Proceedings of the ACM Web Conference 2024},
  pages={3367--3378},
  year={2024}
}

@article{
xu2026asurveyofmodelarchitectures,
title={A Survey of Model Architectures in Information Retrieval},
author={Zhichao Xu and Fengran Mo and Zhiqi Huang and Crystina Zhang and Puxuan Yu and Bei Wang Phillips and Jimmy Lin and Vivek Srikumar},
journal={Transactions on Machine Learning Research},
issn={2835-8856},
year={2026},
url={https://openreview.net/forum?id=xAIbTbHRrX},
note={Survey Certification}
}

@misc{xu2025rethinkingonpolicyoptimizationquery,
      title={Rethinking On-policy Optimization for Query Augmentation}, 
      author={Zhichao Xu and Shengyao Zhuang and Xueguang Ma and Bingsen Chen and Yijun Tian and Fengran Mo and Jie Cao and Vivek Srikumar},
      year={2025},
      eprint={2510.17139},
      archivePrefix={arXiv},
      primaryClass={cs.CL},
      url={https://arxiv.org/abs/2510.17139}, 
}

@article{qian2025toolrl,
  title={Toolrl: Reward is all tool learning needs},
  author={Qian, Cheng and Acikgoz, Emre Can and He, Qi and Wang, Hongru and Chen, Xiusi and Hakkani-T{\"u}r, Dilek and Tur, Gokhan and Ji, Heng},
  journal={arXiv preprint arXiv:2504.13958},
  year={2025}
}

@misc{oord2019representationlearningcontrastivepredictive,
      title={Representation Learning with Contrastive Predictive Coding}, 
      author={Aaron van den Oord and Yazhe Li and Oriol Vinyals},
      year={2019},
      eprint={1807.03748},
      archivePrefix={arXiv},
      primaryClass={cs.LG},
      url={https://arxiv.org/abs/1807.03748}, 
}

@InProceedings{gutmann10noisecontrastiveestimation,
  title = 	 {Noise-contrastive estimation: A new estimation principle for unnormalized statistical models},
  author = 	 {Gutmann, Michael and Hyvärinen, Aapo},
  booktitle = 	 {Proceedings of the Thirteenth International Conference on Artificial Intelligence and Statistics},
  pages = 	 {297--304},
  year = 	 {2010},
  editor = 	 {Teh, Yee Whye and Titterington, Mike},
  volume = 	 {9},
  series = 	 {Proceedings of Machine Learning Research},
  address = 	 {Chia Laguna Resort, Sardinia, Italy},
  month = 	 {13--15 May},
  publisher =    {PMLR},
  url = 	 {https://proceedings.mlr.press/v9/gutmann10a.html}
}

@inproceedings{ma-collins-2018-noise,
    title = "Noise Contrastive Estimation and Negative Sampling for Conditional Models: Consistency and Statistical Efficiency",
    author = "Ma, Zhuang  and
      Collins, Michael",
    editor = "Riloff, Ellen  and
      Chiang, David  and
      Hockenmaier, Julia  and
      Tsujii, Jun{'}ichi",
    booktitle = "Proceedings of the 2018 Conference on Empirical Methods in Natural Language Processing",
    month = oct # "-" # nov,
    year = "2018",
    address = "Brussels, Belgium",
    publisher = "Association for Computational Linguistics",
    url = "https://aclanthology.org/D18-1405/",
    doi = "10.18653/v1/D18-1405",
    pages = "3698--3707"
}

@inproceedings{zhang-stratos-2021-understanding,
    title = "Understanding Hard Negatives in Noise Contrastive Estimation",
    author = "Zhang, Wenzheng  and
      Stratos, Karl",
    editor = "Toutanova, Kristina  and
      Rumshisky, Anna  and
      Zettlemoyer, Luke  and
      Hakkani-Tur, Dilek  and
      Beltagy, Iz  and
      Bethard, Steven  and
      Cotterell, Ryan  and
      Chakraborty, Tanmoy  and
      Zhou, Yichao",
    booktitle = "Proceedings of the 2021 Conference of the North American Chapter of the Association for Computational Linguistics: Human Language Technologies",
    month = jun,
    year = "2021",
    address = "Online",
    publisher = "Association for Computational Linguistics",
    url = "https://aclanthology.org/2021.naacl-main.86/",
    doi = "10.18653/v1/2021.naacl-main.86",
    pages = "1090--1101"
}

@inproceedings{xu-etal-2025-distillation,
    title = "Distillation versus Contrastive Learning: How to Train Your Rerankers",
    author = "Xu, Zhichao  and
      Huang, Zhiqi  and
      Zhuang, Shengyao  and
      Srikumar, Vivek",
    editor = "Inui, Kentaro  and
      Sakti, Sakriani  and
      Wang, Haofen  and
      Wong, Derek F.  and
      Bhattacharyya, Pushpak  and
      Banerjee, Biplab  and
      Ekbal, Asif  and
      Chakraborty, Tanmoy  and
      Singh, Dhirendra Pratap",
    booktitle = "Proceedings of the 14th International Joint Conference on Natural Language Processing and the 4th Conference of the Asia-Pacific Chapter of the Association for Computational Linguistics",
    month = dec,
    year = "2025",
    address = "Mumbai, India",
    publisher = "The Asian Federation of Natural Language Processing and The Association for Computational Linguistics",
    url = "https://aclanthology.org/2025.findings-ijcnlp.33/",
    pages = "564--578",
    ISBN = "979-8-89176-303-6"
}

@inproceedings{ding2024reasoning,
  title={Reasoning and planning with large language models in code development},
  author={Ding, Hao and Fan, Ziwei and Guehring, Ingo and Gupta, Gaurav and Ha, Wooseok and Huan, Jun and Liu, Linbo and Omidvar-Tehrani, Behrooz and Wang, Shiqi and Zhou, Hao},
  booktitle={Proceedings of the 30th ACM SIGKDD Conference on Knowledge Discovery and Data Mining},
  pages={6480--6490},
  year={2024}
}

@article{hu2025qualityflow,
  title={Qualityflow: An agentic workflow for program synthesis controlled by llm quality checks},
  author={Hu, Yaojie and Zhou, Qiang and Chen, Qihong and Li, Xiaopeng and Liu, Linbo and Zhang, Dejiao and Kachroo, Amit and Oz, Talha and Tripp, Omer},
  journal={arXiv preprint arXiv:2501.17167},
  year={2025}
}

@article{Forouzandeh2025MACLA,
  title={Learning Hierarchical Procedural Memory for LLM Agents through
         Bayesian Selection and Contrastive Refinement},
  author={Forouzandeh, Saman and Peng, Wei and Moradi, Parham and
          Yu, Xinghuo and Jalili, Mahdi},
  year={2025},
  month={December},
  day={22},
  eprint={2512.18950},
  archivePrefix={arXiv},
  primaryClass={cs.LG},
  doi={10.48550/arXiv.2512.18950},
  note={Accepted at AAMAS 2026}
}

@article{Yu2026SelfConsolidation,
  title={Self-Consolidation for Self-Evolving Agents},
  author={Yu, Hongzhuo and Zhu, Fei and Xie, Guo-Sen and Shao, Ling},
  year={2026},
  eprint={2602.01966},
  archivePrefix={arXiv},
  primaryClass={cs.LG},
  doi={10.48550/arXiv.2602.01966}
}

@article{Allard2026ERL,
  title={Experiential Reflective Learning for Self-Improving LLM Agents},
  author={Allard, Marc-Antoine and Teinturier, Arnaud and Xing, Victor and Viaud, Gautier},
  journal={arXiv preprint arXiv:2603.24639},
  year={2026},
  eprint={2603.24639},
  archivePrefix={arXiv},
  primaryClass={cs.LG},
  doi={10.48550/arXiv.2603.24639},
  note={ICLR 2026 MemAgents Workshop}
}

@inproceedings{opsahl2024optimizing,
  title={Optimizing instructions and demonstrations for multi-stage language model programs},
  author={Opsahl-Ong, Krista and Ryan, Michael J and Purtell, Josh and Broman, David and Potts, Christopher and Zaharia, Matei and Khattab, Omar},
  booktitle={Proceedings of the 2024 Conference on Empirical Methods in Natural Language Processing},
  pages={9340--9366},
  year={2024}
}

\appendix
\section{Ablation Study}
\label{app:ablation}

We perform some ablation study on \ours\ framework in this section. We will show that all three phases in \ours\ are necessary and removing any of them might result in suboptimal performance of CAM. All the following experiments are conducted on AppWolrd using \cmd{Claude-Sonnet-4} for $\pi^E$.

First, we show that the RL phase in \ours\ is necessary. We remove RL phase and use $\pi^C_{\text{SFT}}$ without RL as CAM. We report the performance of $\pi^C_{\text{SFT}}$ in experiment 3 in \Cref{tab:ablation}. Compared with a full \ours\ framework with SFT + RL (experiment 4), $\pi^C_{\text{SFT}}$ significantly degrade all metrics in TGC and SGC.

Next, we show that contrastive learning (CL) is necessary. To illustrate this, we curate an SFT training dataset $\mc{D}_{\text{SFT\_no\_CL}}$ using only one trajectory for each task. We then perform SFT using $\mc{D}_{\text{SFT\_no\_CL}}$ for CAM and report the performance in experiment 2 in \Cref{tab:ablation}. Compared with SFT using CL (experiment 3), SFT without CL significantly underperforms experiment 3, which shows that CL plays an important role in increasing SFT data quality.

Finally, we show that SFT is necessary. We use a \cmd{Qwen/Qwen3-32B} model directly downloaded from HuggingFace as CAM without any fine-tuning and report the performance in experiment 1 in \Cref{tab:ablation}. Comparing experiment 1 and 3, we see that using data from CL to SFT a CAM has significant performance gain over using an untuned \cmd{Qwen/Qwen3-32B} model as CAM.

\begin{table*}[t]
\centering
\begin{tabular}{c l cc cc}
\toprule
\multirow{2}{*}{Experiment ID} & \multirow{2}{*}{Method} 
& \multicolumn{2}{c}{TGC $\uparrow$} 
& \multicolumn{2}{c}{SGC $\uparrow$} \\
\cmidrule(lr){3-4} \cmidrule(lr){5-6}
& & Avg & Pass@3 & Avg & Pass@3 \\
\midrule
% 0 & Baseline & 72.62(2.59) & 86.90 & 52.38(2.73) & 66.07  \\
1 &  \cmd{Qwen/Qwen3-32B} & 67.40(2.95) & 83.33 & 48.81(4.49) & 66.07\\
2& ~~ SFT without CL &  68.65 (2.68) & 88.69 & 42.86 (4.72) & 67.86\\
3& ~~ SFT with CL & 74.21(1.72) & 90.48 & 50.60(2.73) & 71.43\\
4& \ours & \textbf{81.15}(2.48) & \textbf{91.67} & \textbf{66.67}(4.49) & \textbf{82.14} \\
\bottomrule
\end{tabular}
\caption{Ablation study on \textbf{AppWorld} \textsc{Test-N} split. For the CAM, we use the following variant: (1) a \cmd{Qwen/Qwen3-32B} model without SFT and RL as in experiment 1. (2) a Qwen3-32B model after SFT on $\mc{D}_{\text{SFT\_no\_CL}}$ (no RL) as in experiment 2. (3) a Qwen3-32B model after SFT with CL (no RL) as in experiment 3. (4) full \ours\ framework as in experiment 4.
Results are averaged over three runs (standard deviation in parentheses) except the Pass@3 metric.}
\label{tab:ablation}
\end{table*}
\section{Latency Study}
\label{app:latency}

In this section, we present latency study for triggering a CAM. We report the averaged task execution time, averaged number of turns of the execution agent $\pi^E$, averaged throughput of CAM, and averaged latency for invoking CAM. The average is taken across AppWorld \textsc{Test-N} split. The CAM is hosted via vllm on 8 NVIDIA B200 GPUs. 

\begin{table*}[t]
\centering
\begin{tabular}{l cccc}
\toprule
 Method &  Run time (s) & \# of turns & CAM throughput (tokens/s) & CAM latency (s)\\
\midrule
Baseline & 63.4 & 17.3 & NA & NA \\
\ours~(ours) & 77.2 & 18.5 & 69.5 & 1.21\\
\bottomrule
\end{tabular}
\caption{CAM Latency study. Task run time, number of turns of the execution agent $\pi^E$, throughput of CAM, and latency for invoking CAM. All numbers are averaged over AppWorld \textsc{Test-N} split. The model for $\pi^E$ is \cmd{Claude-Sonnet-4}.}
\label{tab:latency}
\end{table*}

From \Cref{tab:latency}, we see that on average, \ours\ increases number of turns by 1.2 and triggers a latency of 1.2 second to invoke CAM. These two components sum up to a 13.8 second increase in terms of task run time. Overall, this additional overhead is modest compared to the performance gains achieved by \ours.
\section{Comparison to RAG}
\label{app:rag}

We discuss the similarities and differences between \ours\ and RAG. 
Our augmentation model $\pi^C(\cdot)$ can be effectively viewed as a \emph{generative retrieval} model, following the idea of generate-then-read~\citep{yu2022generate}. 
For each task, it \emph{generates} the most useful context from its internal parameters, rather than \emph{retrieving} the most similar context from an external knowledge base.

The key difference lies in how the retrieved information is used. 
In RAG, knowledge items are retrieved as-is, and the execution agent must reason over them to determine how they should be applied to the current task, assuming the current task is not available in knowledge base. This is true because the knowledge base is established using the training set, which is disjoint from the test set.
In contrast, \ours\ shifts this reasoning burden to the augmentation model $\pi^C$, which generates context that is already tailored to the new query. 
As a result, the generated context $c$ is directly actionable for the execution agent $\pi^E$, reducing the amount of additional reasoning required from $\pi^E$, particularly when the execution model is relatively weak. 

This statement is supported by the experiments with \cmd{DeepSeek-V3.1} on AppWorld, as shown in \Cref{tab:deepseek}. 
Compared with \cmd{Claude-Sonnet-4}, \cmd{DeepSeek-V3.1} is generally considered less capable (see their respective model cards). 
Under this setting, the RAG baseline with \cmd{DeepSeek-V3.1} even underperforms the vanilla baseline, suggesting that simply retrieving items by embedding similarity is noisy when the underlying model lacks strong reasoning ability. 
In contrast, \ours\ improves performance by generating task-specific context that is already adapted to the new query. Similar performance degradation can be observed for ACE, whose playbook is curated from training trajectories. As a result, the execution agent $\pi^E$ must still perform additional reasoning to determine how the retrieved instructions apply to the current task.
\section{CAM Transferability}
\label{app:transferability}
The objective of CAM is to provide auxiliary context and can be detached from the task execution agent $\pi^E$. In this section, we study whether a trained CAM can be applied to a different $\pi^E$ without retraining.

Recall that the training dataset $\mathcal{D}_{\text{train}}$ is generated by contrastive analysis of the replay buffer $\Gamma$, which is generated by the execution agent $\pi^E$ powered by \cmd{Claude-Sonnet-4}. 
The reflection agent $\pi^R$ that analyzes $\Gamma$ is also powered by \cmd{Claude-Sonnet-4}. 
Furthermore, during Phase~2 RL training, the execution agent $\pi^E$ used for reward computation also uses \cmd{Claude-Sonnet-4} as its foundation model. 
Consequently, the entire CAM training pipeline relies solely on trajectories and feedback generated by the Claude model.

Despite this, the trained CAM demonstrates strong transferability. 
At test time, it still provides significant performance gains when the execution agent $\pi^E$ is replaced by a different model, such as \cmd{DeepSeek-V3.1}. For example, averaged TGC and SGC increase by $1.68$ and $5.35$ over the baseline respectively, as shown in \Cref{tab:deepseek}. 
This result suggests that once trained with the \ours\ framework, the CAM can generalize across different execution agents without requiring retraining.

\begin{table*}[t]
\centering

\begin{tabular}{l l cc cc}
\toprule
\multirow{2}{*}{Model for $\pi^E$} & \multirow{2}{*}{Method} 
& \multicolumn{2}{c}{TGC $\uparrow$} 
& \multicolumn{2}{c}{SGC $\uparrow$} \\
\cmidrule(lr){3-4} \cmidrule(lr){5-6}
& & Avg & Pass@3 & Avg & Pass@3 \\
\midrule

% \multirow{4}{*}{\cmd{Claude-Sonnet-4}}
% & Baseline & 72.62(2.59) & 86.90 & 52.38(2.73) & 66.07  \\
% & RAG & 72.02(2.15) & 86.31 & 54.67(3.72) & 71.43  \\
% % & \textsc{GenRead} & 64.60(1.92) & 73.21 & 44.84(3.28) & 57.14\\
% & ACE
% % ~\cite{zhang2025agenticcontextengineeringevolving} 
% & 74.40(3.57) & 85.71 & 58.93(6.19) & 73.21 \\
% & \ours~(ours) & \textbf{81.15}(2.48) & \textbf{91.67} & \textbf{66.67}(4.49) & \textbf{82.14} \\

% \cmidrule(l){1-6}

\multirow{4}{*}{\cmd{DeepSeek-V3.1}}
& Baseline &  41.27(3.05) & 64.29 & 19.05(2.06) & 26.79  \\
& RAG & 33.33(1.57) & 53.57 & 10.71(4.72) & 21.43  \\
% & \textsc{GenRead} & 26.98(2.68) & 42.26 & 13.69(2.73) & 23.21\\
& ACE
% ~\cite{zhang2025agenticcontextengineeringevolving} 
& 32.54(1.50) & 53.57 & 13.69(1.03) & 28.57\\
& \ours~(ours) & \textbf{42.95}(2.85) & \textbf{66.07} & \textbf{24.40}(4.49) & \textbf{33.93}  \\

\bottomrule
\end{tabular}
\caption{\textbf{AppWorld} experiments results. Task Goal Completion (TGC) and Scenario Goal Completion (SGC) on the \textsc{Test-N} split are reported. Results are averaged over three runs (standard deviation in parentheses) except the Pass@3 metric.}
\label{tab:deepseek}

\end{table*}
\section{Experiment Setting}
\label{app:training}

We introduce our experiment setting for all three phases: agentic reflection, SFT, and RL.

\subsection{Agentic Data Collection}

\paragraph{Reflection Agent.}
We use the Strands Agents framework together with \cmd{Claude-Sonnet-4} to build a reflection agent for contrastive analysis. The full prompt used for the reflection agent is provided in \Cref{app:prompt}.

If all $m=6$ runs of a task are processed by a single reflection pass, the resulting dataset $\mathcal{D}_{\text{SFT}}$ would have the same size as the training dataset $\mc{D}_{\text{train}}$, which is too small to effectively fine-tune LLMs with billions of parameters.

To increase the amount of data, for each task we sample subsets of 3 runs from the 6 collected trajectories. 
The reflection agent only analyzes the 3 selected runs. This process can be repeated for $\binom63 = 20$ times, which effectively expands the training dataset by a factor of $20$.
% and results in 1{,}798 training samples as $\mc{D}_{\text{train}}$.

\subsection{Supervised Fine-Tuning}

We further randomly split $\mathcal{D}_{\text{SFT}}$ into 80\% for training and 20\% for validation. 
The CAM $\pi^C_\theta(\cdot)$ is initialized from a \cmd{Qwen/Qwen3-32B} model~\citep{yang2025qwen3}, downloaded from HuggingFace\footnoteurl{https://huggingface.co/Qwen/Qwen3-32B}. 
We perform full-parameter fine-tuning for 5 epochs using 8 NVIDIA B200 GPUs. 
Training is conducted in bf16 precision with a learning rate of $1\times10^{-5}$ and a warm-up ratio of 0.05. The supervised fine-tuning is implemented using the \textsc{LlamaFactory} framework~\citep{zheng2024llamafactory}.

\subsection{Reinforcement Learning with GRPO}

We perform reinforcement learning on the augmentation model $\pi^C_{\text{SFT}}$ using the GRPO algorithm, implemented with the Verl framework~\citep{sheng2024hybridflow}. 
Training is conducted for 15 epochs on the train dataset using 8 NVIDIA B200 GPUs.

As described in \Cref{subsec:training}, computing the GRPO reward requires multi-turn interactions between the task execution agent $\pi^E_\theta$ and the benchmark environment ${M}$ defined in \Cref{eq:pomdp}. 
To efficiently compute rewards in batch, we also leverage Amazon Bedrock AgentCore Runtime, which bootstraps multiple containers in parallel to support high-concurrency rollout execution hence reward computation.
Additional hyperparameter configurations for GRPO are provided below.

\begin{lstlisting}[language=bash]
python3 -m verl.trainer.main_ppo \
    algorithm.adv_estimator=grpo \
    data.train_files=<path/to/train.parquet> \
    data.val_files=<path/to/train.parquet> \
    data.train_batch_size=4 \
    data.max_prompt_length=4096 \
    data.max_response_length=1024 \
    data.filter_overlong_prompts=True \
    data.truncation='error' \
    actor_rollout_ref.model.path=<model_path> \
    actor_rollout_ref.actor.optim.lr=1e-6 \
    actor_rollout_ref.model.use_remove_padding=True \
    actor_rollout_ref.actor.ppo_mini_batch_size=2 \
    actor_rollout_ref.actor.ppo_micro_batch_size_per_gpu=1 \
    actor_rollout_ref.actor.use_kl_loss=True \
    actor_rollout_ref.actor.kl_loss_coef=0.001 \
    actor_rollout_ref.actor.kl_loss_type=low_var_kl \
    actor_rollout_ref.actor.entropy_coeff=0 \
    actor_rollout_ref.model.enable_gradient_checkpointing=True \
    actor_rollout_ref.actor.fsdp_config.param_offload=False \
    actor_rollout_ref.actor.fsdp_config.optimizer_offload=False \
    actor_rollout_ref.rollout.log_prob_micro_batch_size_per_gpu=2 \
    actor_rollout_ref.rollout.tensor_model_parallel_size=2 \
    actor_rollout_ref.rollout.name=vllm \
    actor_rollout_ref.rollout.gpu_memory_utilization=0.5 \
    actor_rollout_ref.rollout.n=4 \
    actor_rollout_ref.ref.log_prob_micro_batch_size_per_gpu=2 \
    actor_rollout_ref.ref.fsdp_config.param_offload=True \
    algorithm.use_kl_in_reward=False \
    trainer.critic_warmup=0 \
    trainer.n_gpus_per_node=8 \
    trainer.nnodes=1 \
    trainer.save_freq=20 \
    trainer.test_freq=20 \
    trainer.total_epochs=15 \
    actor_rollout_ref.rollout.checkpoint_engine.update_weights_bucket_megabytes=4096 \
    custom_reward_function.path=<reward_function_path> $@
\end{lstlisting}

\section{Prompt for Reflection Agent}
\label{app:prompt}
% \red{@Guande, Yawei, Han: Need review}

We provide the system prompt and user/task prompt for the reflection agent $\pi^R(\cdot)$. They are universal across different benchmarks.
\subsection{System Prompt}
\begin{tcblisting}{
  colback=gray!5,
  colframe=gray!40,
  title={System Prompt for the Reflection Agent},
  fonttitle=\bfseries,
  listing only,
  breakable,
  enhanced,
  boxrule=0.6pt,
  left=1mm,right=1mm,top=1mm,bottom=1mm,
}
Your task is to analyze LLM agent execution trajectories and generate guidance using contrastive learning.

## CRITICAL: Context Window Management - READ THIS FIRST

**WARNING**: Trajectory files can be EXTREMELY large (100K-500K+ tokens). Reading entire files will cause out-of-context-window errors and crash the session.

**MANDATORY FIRST STEP**: Before reading ANY file content, you MUST check its size:
```bash
# ALWAYS run this first for any file
ls -lh <file_path>
wc -l <file_path>  # Count lines
wc -c <file_path>  # Count bytes
```

**Size Guidelines**:
- Files < 10KB: Safe to read with `cat` or `jq`
- Files 10KB-100KB: Use `head`, `tail`, or targeted `jq` queries
- Files > 100KB: ONLY use regex/grep patterns, NEVER read full content
- Files > 1MB: Use aggressive filtering with grep + head limits

**NEVER DO THIS**:
- `cat large_file.json` - Will overflow context
- `jq '.' large_file.json` - Will overflow context
- Reading entire message arrays without limits

**ALWAYS DO THIS**:
- Check file size FIRST
- Use regex patterns with grep to search
- Limit output with `head -n` or `tail -n`
- Use `jq` with specific field selectors, not full dumps

## Available Tools

You have access to the `shell` tool to execute bash commands.

### Step 1: ALWAYS Check File Sizes First
```bash
# MANDATORY: Run these commands BEFORE any content extraction
ls -lh <trajectory_folder>/*.json
du -sh <trajectory_folder>

# List files sorted by size
ls -lhS <trajectory_folder>/*.json | head -20
```

### Step 2: Use Regex Patterns for Searching (Primary Method)
```bash
# Search for error patterns using regex - MOST EFFICIENT
grep -E "error|Error|ERROR|exception|Exception
|failed|Failed" <file_path> | head -30
grep -oE '"error":\\s*"[^"]*"' <file_path> | head -20
grep -oE '"success":\\s*(true|false)' <file_path> | head -10

# Search for specific patterns with context
grep -E -B2 -A2 "error|exception" <file_path> | head -50

# Count occurrences of patterns
grep -cE "error|failed|exception" <file_path>

# Find tool calls and their patterns
grep -oE '"tool_name":\\s*"[^"]*"' <file_path> | sort | uniq -c | sort -rn

# Extract specific JSON values with regex
grep -oE '"reward":\\s*[0-9.-]+' <file_path>
grep -oE '"task_id":\\s*"[^"]*"' <file_path>

# Search for repeated patterns (loops, stuck behavior)
grep -oE '"content":\\s*"[^"]{0,100}' <file_path> | sort | uniq -c | sort -rn | head -20
```

### Step 3: Targeted jq Queries (Only for Small Extractions)
```bash
# ONLY use jq for small, targeted extractions
jq 'keys' <file_path>  # Structure overview (safe)
jq '.data.messages | length' <file_path>  # Count messages (safe)
jq '.data.messages[0]' <file_path>  # First message only
jq '.data.messages[-1]' <file_path>  # Last message only
jq '{reward: .data.reward, success: .data.metrics.success}' <file_path>  # Summary stats

# DANGEROUS - Only use with head limit:
jq -r '.data.messages[].content // empty' <file_path> 2>/dev/null | head -50
```

## Contrastive Learning Strategy

1. **Identify successful vs failed traces** using reward/success metrics
2. **Compare agent behaviors** between successful and failed cases
3. **Find patterns** that differentiate success from failure
4. **Generate insights** about what works and what doesn't

## Output Format

Your final output should be a task-specific guidance in the following format:
<guidance>
The strategy for completing this task is ...
</guidance>
\end{tcblisting}

\subsection{User Prompt}
\begin{tcblisting}{
  colback=gray!5,
  colframe=gray!40,
  title={User Prompt for the Reflection Agent},
  fonttitle=\bfseries,
  listing only,
  breakable,
  enhanced,
  boxrule=0.6pt,
  left=1mm,right=1mm,top=1mm,bottom=1mm,
}
You are given a folder that contains multiple LLM agent trajectory traces for solving the same task. 
They are a mix of successful and failed executions in general, indicated by reward/success metrics in the traces, but it's also possible that all traces are successful or failed.
Please analyze multiple LLM agent trajectories and generate a guidance using contrastive learning.

## Trajectory Folder
Path: {{ traces_folder }}

This folder contains multiple trajectory JSON files - a mix of successful and failed agent executions. You can work outside this folder.

## CRITICAL: Instructions (Follow in Order)

**Step 1: CHECK FILE SIZES FIRST (MANDATORY)**
Before reading ANY trajectory content, run:
```bash
ls -lh {{ traces_folder }}/*.json 2>/dev/null || ls -lh {{ traces_folder }}
find {{ traces_folder }} -name "*.json" -exec ls -lh {} \; | head -20
```
This prevents context window overflow errors.

**Step 2: IDENTIFY SUCCESS VS FAILURE (CONTRASTIVE SETUP)**
Extract reward/success metrics from all files to categorize them:
```bash
# Get reward values from all trajectory files
for f in {{ traces_folder }}/*.json; do
    echo -n "$f: "
    grep -oE '"reward":\s*[0-9.-]+' "$f" | head -1
done 2>/dev/null | head -30

# Or use jq if files are small
find {{ traces_folder }} -name "*.json" -size -100k -exec sh -c 'echo -n "{}: " && jq -r ".data.reward // .reward // \"N/A\"" "{}"' \; 2>/dev/null | head -30
```

Categorize traces into:
- **Successful traces**: High reward (e.g., reward > 0.5 or success=true)
- **Failed traces**: Low reward (e.g., reward <= 0 or success=false)

**Step 3: ANALYZE SUCCESSFUL TRACES**
For successful traces, identify:
- What actions/strategies led to success
- How the agent handled challenges
- Decision-making patterns
- Skip this step if there is no successful trace

```bash
# Example: Analyze a successful trace (adjust filename)
grep -oE '"tool_name":\s*"[^"]*"' <successful_trace.json> | sort | uniq -c | sort -rn
grep -E -B2 -A2 "success|complete|done" <successful_trace.json> | head -50
```

**Step 4: ANALYZE FAILED TRACES**
For failed traces, identify:
- What went wrong (errors, loops, wrong decisions)
- Where the agent got stuck
- Common failure patterns
- Skip this step if there is no failed traces

```bash
# Example: Analyze a failed trace (adjust filename)
grep -E "error|Error|failed|Failed|exception" <failed_trace.json> | head -30
grep -oE '"content":\s*"[^"]{0,100}' <failed_trace.json> | sort | uniq -c | sort -rn | head -20
```

**Step 5: CONTRASTIVE COMPARISON**
Compare successful vs failed traces to identify:
1. **Differentiating behaviors**: What do successful agents do that failed ones don't?
2. **Common pitfalls**: What mistakes do failed agents make?
3. **Recovery strategies**: How do successful agents recover from errors?
4. **Decision quality**: How do successful agents make better choices?

**Step 6: GENERATE TASK-SPECIFIC GUIDANCE**
Based on your contrastive analysis, generate a step-by-step guidance for this task. The goal is to add this guidance at the beginning of the task, so that the agent can always successfully complete the task. Be concise.

## Output Requirements

After your analysis, output the task-specific guidance in the following format:

<guidance>
The strategy for completing this task is ...
</guidance>
\end{tcblisting}

\section{Prompt for AppWorld}
For AppWorld dataset, We use the official system prompt released at \url{https://github.com/StonyBrookNLP/appworld/blob/main/experiments/prompts/react_code_agent/_legacy_instructions.txt} for $\pi^E$. For completeness, we include it below.

\begin{tcblisting}{
  colback=gray!5,
  colframe=gray!40,
  title={User Prompt for the Reflection Agent},
  fonttitle=\bfseries,
  listing only,
  breakable,
  enhanced,
  boxrule=0.6pt,
  left=1mm,right=1mm,top=1mm,bottom=1mm,
}
USER:
I am your supervisor and you are a super intelligent AI Assistant whose job is to achieve my day-to-day tasks completely autonomously.

To do this, you will need to interact with app/s (e.g., spotify, venmo etc) using their associated APIs on my behalf. For this you will undertake a *multi-step conversation* using a python REPL environment. That is, you will write the python code and the environment will execute it and show you the result, based on which, you will write python code for the next step and so on, until you've achieved the goal. This environment will let you interact with app/s using their associated APIs on my behalf.

Here are three key APIs that you need to know to get more information

# To get a list of apps that are available to you.

```python
print(apis.api_docs.show_app_descriptions())
```

# To get the list of apis under any app listed above, e.g. spotify

```python
print(apis.api_docs.show_api_descriptions(
app_name='spotify'))
```

# To get the specification of a particular api, e.g. spotify app's login api

```python
print(apis.api_docs.show_api_doc(
app_name='spotify', api_name='login'))
```

Each code execution will produce an output that you can use in subsequent calls. Using these APIs, you can now generate code, that I will execute, to solve the task. Let's start with the task

<An in-context learning example. Omitted for better readability.>

----------------------------------------------

USER:
**Key instructions**:
(1) Make sure to end code blocks with ``` followed by a newline(\n).

(2) Remember you can use the variables in your code in subsequent code blocks.

(3) Remember that the email addresses, access tokens and variables (e.g. spotify_password) in the example above are not valid anymore.

(4) You can use the "supervisor" app to get information about my accounts and use the "phone" app to get information about friends and family.

(5) Always look at API specifications (using apis.api_docs.show_api_doc) before calling an API.

(6) Write small chunks of code and only one chunk of code in every step. Make sure everything is working correctly before making any irreversible change.

(7) Many APIs return items in "pages". Make sure to run through all the pages by looping over `page_index`.

(8) Once you have completed the task, make sure to call apis.supervisor.complete_task(). If the task asked for some information, return it as the answer argument, i.e. call apis.supervisor.complete_task(answer=
<answer>). Many tasks do not require an answer, so in those cases, just call apis.supervisor.complete_task() i.e. do not pass any argument.

USER:
Using these APIs, now generate code to solve the actual task:

My name is: {{ main_user.first_name }} {{ main_user.last_name }}. My personal email is {{ main_user.email }} and phone number is {{ main_user.phone_number }}.
Task: {{ input_str }}
\end{tcblisting}

\section{WebShop Prompt}
We provide the systemp prompt of the execution agent $\pi^E(\cdot)$ for WebShop.

\begin{tcblisting}{
  colback=gray!5,
  colframe=gray!40,
  title={User Prompt for the Reflection Agent},
  fonttitle=\bfseries,
  listing only,
  breakable,
  enhanced,
  boxrule=0.6pt,
  left=1mm,right=1mm,top=1mm,bottom=1mm,
}
You are an expert online shopping assistant designed to help users find and purchase products that match their specific requirements on an e-commerce website.

Your goal is to navigate the shopping website efficiently and complete purchase tasks by:
1. Understanding the user's product requirements completely
2. Searching for relevant products using appropriate keywords
3. Evaluating product details to ensure they match ALL requirements
4. Selecting the correct product options (size, color, etc.)
5. Completing the purchase by clicking "buy now"

## Available Tools

You have access to three tools to interact with the website:

### 1. search(query: str)
Search for products using space-separated keywords.
- Keywords are matched against product titles, descriptions, and attributes
- Use specific keywords that capture the essential requirements
- Examples: "red shirt large", "wooden toy", "laptop bag waterproof"

### 2. click(element: str)
Click on elements like product links, buttons, or options.
- Product ASINs (e.g., "b09mnnbdtj") - to view product details
- Navigation: "next >", "< prev", "back to search"
- Options: "red", "blue", "small", "medium", "large", etc.
- Tabs: "description", "features", "reviews", "attributes"
- Action: "buy now" - to complete purchase (USE THIS WHEN READY!)

### 3. get_available_actions()
Get a list of all clickable elements on the current page.
- Use this when you're unsure what actions are available
- Helpful when clicks are failing or you need to see all options
- Returns: has_search_bar boolean and list of clickable elements

## Task Completion Strategy

1. **Understand Requirements**: Carefully read the task to identify all product requirements (type, color, size, features, etc.)

2. **Search Effectively**:
    - Use relevant keywords that match the product requirements
    - Start with general terms, then refine if needed
    - Review search results to find promising products

3. **Evaluate Products**:
    - Click on product ASINs to view details
    - Check description, features, and attributes tabs
    - Verify the product matches ALL requirements from the task

4. **Select Options**:
    - If the product has options (size, color), select the ones matching requirements
    - Use get_available_actions() if you need to see what options are available

5. **Complete Purchase**:
    - Once you've found a product that matches all requirements and selected appropriate options
    - Click "buy now" to complete the task
    - This will end the episode and you'll receive a reward

## Important Tips

- Be thorough: Verify all requirements before purchasing
- If search results don't match, try different keywords
- Use get_available_actions() when uncertain about next steps
- Don't buy products that only partially match requirements
- Read product details carefully - don't assume based on title alone

## Example Workflow

Task: "Find a red shirt in size large"

1. search("red shirt large")
2. Click on promising product ASIN
3. Check description/features to verify it's actually red
4. Look for size options and click("large")
5. Verify all requirements are met
6. click("buy now")

Now let's complete your shopping task!
\end{tcblisting}

\end{document}